\documentclass{article}

\usepackage{arxiv}

\usepackage{amsmath,amsfonts,bm}

\def\eqref#1{equation~\ref{#1}}

\def\1{\bm{1}}

\def\vf{{\bm{f}}}
\def\vg{{\bm{g}}}

\def\vp{{\bm{p}}}
\def\vq{{\bm{q}}}

\def\vs{{\bm{s}}}

\def\vu{{\bm{u}}}

\def\mA{{\bm{A}}}

\def\mK{{\bm{K}}}

\DeclareMathAlphabet{\mathsfit}{\encodingdefault}{\sfdefault}{m}{sl}
\SetMathAlphabet{\mathsfit}{bold}{\encodingdefault}{\sfdefault}{bx}{n}

\def\sR{{\mathbb{R}}}

\usepackage{hyperref}
\hypersetup{
    colorlinks=true,
    citecolor=blue
}

\usepackage{url}
\usepackage{booktabs}
\usepackage{graphicx}
\usepackage{algorithm} 
\usepackage{algpseudocode}
\usepackage{varwidth}
\usepackage{multicol}
\usepackage{multirow}
\usepackage{caption}
\usepackage{subcaption}
\usepackage{wrapfig}

\usepackage[utf8]{inputenc} 
\usepackage[T1]{fontenc}    
\usepackage{hyperref}       
\usepackage{url}            
\usepackage{booktabs}       
\usepackage{amsfonts}       
\usepackage{nicefrac}       
\usepackage{microtype}      
\usepackage{lipsum}
\usepackage{amssymb}
\usepackage{pifont}
\usepackage{xcolor}
\usepackage{longtable}
\usepackage{diagbox}

\usepackage{colortbl}
\usepackage{pgf}
\usepackage{zref-savepos}

\title{When Physics Meets Machine Learning: A Survey of Physics-Informed Machine Learning}

\author{
  Chuizheng Meng, Sungyong Seo, Defu Cao, Sam Griesemer, Yan Liu\\
  Department of Computer Science\\
  University of Southern California\\
  Los Angeles, CA 90089 \\
  \texttt{\{chuizhem, sungyongs, defucao, samgriesemer, yanliu.cs\}@usc.edu} \\
}

\begin{document}
\maketitle

\begin{abstract}
Physics-informed machine learning (PIML), referring to the combination of prior knowledge of physics, which is the high level abstraction of natural phenomenons and human behaviours in the long history, with data-driven machine learning models, has emerged as an effective way to mitigate the shortage of training data, to increase models' generalizability and to ensure the physical plausibility of results. In this paper, we survey an abundant number of recent works in PIML and summarize them from three aspects: (1) motivations of PIML, (2) physics knowledge in PIML, (3) methods of physics knowledge integration in PIML. We also discuss current challenges and corresponding research opportunities in PIML.
\end{abstract}



\section{Introduction}
Machine learning/deep learning models have already achieved tremendous success in a number of domains such as computer vision~\cite{lecun1998gradient,krizhevsky2012imagenet,he2016deep,redmon2016you,he2017mask} and natural language processing~\cite{mikolov2013efficient,socher2013recursive,sutskever2014sequence,kim2014convolutional,bahdanau2014neural,kumar2016ask,vaswani2017attention,peters2018deep,devlin2019bert}, where large amounts of training data and highly expressive neural network architectures together give birth to solutions outperforming previously dominating methods. As a consequence, researchers have also started exploring the possibility of applying machine learning models to advance scientific discovery and to further improve traditional analytical modeling~\cite{hsieh2009machine,ivezic2014statistics,karpatne2017theory,karpatne2018machine,kutz2017deep,reichstein2019deep,wang2018successful}.

While given a set of input and output pairs, deep neural networks are able to extract the complicated relations between the input and output via appropriate optimization over adequate large amount of data, \textit{prior knowledge} still acts as an important role in finding the optimal solution. As the high level extraction of data distributions and task properties, prior knowledge, if incorporated properly, can provide rich information not existing or hard to extract in limited training data, and helps improve the data efficiency, the ability to generalize, and the plausibility of resulting models.

Physics knowledge, which has been collected and validated explicitly both theoretically and empirically in the long history, contains tremendous abstraction and summary of natural phenomena and human behaviours in many important scientific and engineering applications. Thus in this paper, we focus on the topic of integrating prior physics knowledge into machine learning models, i.e. \textbf{\textit{physics-informed machine learning (PIML)}}. Compared to the integration of other types of prior knowledge, such as knowledge graphs, logic rules and human feedback~\cite{von2019informed}, the integration of physics knowledge requires specific design due to its special properties and forms.

In this paper, we survey a wide range of recent works in PIML and summarize them from three aspects. (1) Motivations of PIML, which can be further categorized to using machine learning to serve tasks in physics domains and incorporating physics principles to existing machine learning models for real-world tasks. (2) Physics knowledge in PIML, each type of which is a general principle covering a wide range of problems. (3) Methods of physics knowledge integration in PIML. Depending on the location of knowledge integration, we categorize the methods into data enhancement, neural network architecture design, and physics-informed optimization.

The paper is organized as follows. Sec~\ref{sec:purposes-of-piml} analyzes two main categories of motivations using PIML: one mainly serves tasks in physics domain, while the other serves real-world problems. Sec~\ref{sec:physics-knowledge-in-piml} introduces several general physics principles widely used in PIML. Sec~\ref{sec:methods-of-piml} investigates methods of physics knowledge integration. Sec~\ref{sec:discussion} discusses challenges and potential future research directions of PIML. Sec~\ref{sec:conclusion} serves as the summary of the whole paper.
\section{Motivations of PIML}
\label{sec:purposes-of-piml}
\subsection{ML for Physics: Enhancement of Physics Models via Data-Driven Methods}
Physical science problems involves various data-intensive tasks including spatiotemporal data modeling, causal reasoning, computer vision, probabilistic inference and so on. Since machine learning methods have achieved great success in these tasks, using machine learning models for furthering scientific discovery in physics has received increasing interest in recent years. Compared to existing numerical or pure physics based methods, physics-informed machine learning methods have advantages in flexibility, generalizability, and computation cost. Meanwhile, they still enforce physical plausibility. In this section, we introduce recent developments of exploiting machine learning for several physics related tasks, including surrogate simulation, data-driven PDE solvers, parameterization of physics models, reduced-order models, and knowledge discovery.

\subsubsection{Simulation}

Many physics-informed machine learning models have been proposed to act as surrogate solutions to numerical simulation in many domains, such as turbulence simulation~\cite{esmaeilzadeh2020meshfreeflownet,wang2020towards,kashinath2020enforcing}, climate simulation~\cite{rasp2020weatherbench,weyn2020improving,gronquist2021deep,kashinath2021physics}, and particle system simulation~\cite{sanchez2020learning,li2018learning,ummenhofer2019lagrangian,pfaff2021learning}. Compared to numerical simulation, neural network based machine learning models enjoy following advantages. (1) \textbf{Lower implementation costs.} To build a high-quality numerical simulator, researchers usually need years of engineering effort and must choose numerous physically meaningful parameters depending on the task. Instead, machine learning models can be trained directly from a large amount of observed data. (2) \textbf{Stronger ability to generalize.} Machine learning models can share the same architecture for different types of problems in the same category, and then can be further specialized for each problem with observed data. For example, ~\cite{sanchez2020learning} proposes a general framework to learn particle simulation, and can generalize across fluid, rigid , and deformable material systems. (3) \textbf{Lower computation costs.} Existing neural network building blocks such as multilayer perceptrons and convolutions can be efficiently accelerated by various hardware including CPU, GPU, FPGA and ASIC, giving advantages of computation costs to neural networks composed of these blocks. In fluid flow prediction, the model proposed in ~\cite{belbute2020combining} achieves prediction results close to ground truth simulation at a running speed 60x faster than the numerical method.

Meanwhile, due to the large and complex optimization space of neural networks, it is critical to incorporate inductive bias of physics knowledge about the task into either the training data, the model architecture, or the optimization process. This ensures greater physical plausibility of resulting models, further improving the robustness in real-world settings. The detailed techniques of integration will be introduced in Section~\ref{sec:methods-of-piml}.

\subsubsection{Data-driven PDE Solver} Many real-world problems from physical systems are mostly about how to describe observations based on partial differential equations and numerically solve the equations.
There are many traditional methods to numerically solve PDEs such as spectral methods, finite-difference methods (FDM), finite element method (FEM), finite volume method (FVM), etc.
All of these methods are numerical and they require proper discretization or a finite number of steps to approximate continuous solution.

Combining machine learning techniques with PDE models has a long history in machine learning.
~\cite{crutchfield1987equations} introduce a method to reconstruct the deterministic portion of the equations of motion directly from a data series. 
This approach employs an informational measure of model optimality to guide searching through the space of dynamical systems.
\cite{kevrekidis2003equation} present a framework for computer-aided multi scale analysis, which enables models at a fine (microscopic/stochastic) level of description to perform modeling tasks at a coarse (macroscopic/systems) level.
These macroscopic modeling tasks, yielding information over long time and large scales, are accomplished through approximately initialized calls to the microscopic similar for only short times and small spatial domains.

More recently, \cite{raissi2017physicsI,raissi2017physicsII} introduce a solution to solve PDE in a data-driven manner.
Rather than analytically solving a given equation, they infer solutions to targeted PDE via supervised learning.
Given an input tuple $(x,t)$, they compute spatial and time derivatives from black-box models and then the output are connected based on a form of PDE to update all learnable parameters in the black-box models.
This method does not require any discretization and it is fully data-driven to find a surrogate model.

\cite{raissi2018deep} approximate the unknown solution as well as the nonlinear dynamics by two deep neural networks.
The first network acts as a prior on the unknown solution and essentially enables us to avoid numerical differentiations which are inherently ill-conditioned and unstable.
The second network represents the nonlinear dynamics and helps us distill the mechanisms that govern the evolution of a given spatiotemporal dataset.
\cite{magill2018neural} introduce a technique based on the singular vector canonical correlation analysis (SVCCA) for measuring the generality of neural network layers across a continuously-parameterized set of tasks.
They illustrate this method by studying generality in neural networks trained to solve parameterized boundary value problems based on the Poisson partial differential equation.
\cite{li2020neural} directly model the mapping between a PDE's solution and its initial conditions via message passing in the spatial domain.~ The following work~\cite{li2021fourier} further extends the modeling to the frequency domain, which inherently generalizes to multiple spatial resolutions.

In many physical systems, the governing equations are known with high confidence, but direct numerical solution is prohibitively expensive. 
Often this situation is alleviated by writing effective equations to approximate dynamics below the grid scale.
This process is often impossible to perform analytically and is often ad hoc. \cite{bar2019learning} propose data-driven discretization, a method that uses machine learning to systematically derive discretizations for continuous physical systems.
\cite{um2020sol} target the problem of reducing numerical errors of iterative PDE solvers and compare different learning approaches for finding complex correction functions.
They integrate the PDE solver into the training loop and thereby allow the model to interact with the PDE during training.


\paragraph*{Downscaling}
Directly solving PDEs require spatial and temporal discretization and finer resolution is more desired to capture physically reliable solutions.
However, it increases the computational cost and modeling complexity. 
Downscaling techniques have been widely used as a solution to capture physical variables that
need to be modeled at a finer resolution from a coarser resolution.
Recently, artificial neural networks have
shown a lot of promise for this problem, given their ability to model non-linear relationships.
\cite{sharifi2019downscaling} present a downscaling algorithms using neural networks to leverage the relationships between Satellite precipitation estimates (SPEs) and cloud optical and microphysical properties in northeast Austria. 
\cite{vandal2017deepsd} present DeepSD (Statistically Downscaling), a generalized stacked super resolution convolutional neural network (SRCNN) framework for statistical downscaling of climate variables. DeepSD augments SRCNN with multi-scale input channels to maximize predictability in statistical downscaling. 
\cite{lee2020coarse} introduce a data-driven framework for the identification of unavailable coarse-scale PDEs from microscopic observations via machine-learning algorithms.
Specifically, using Gaussian processes, artificial neural networks, and/or diffusion maps, the proposed framework uncovers the relation between the relevant macroscopic space fields and their time evolution (the right-hand side of the explicitly unavailable macroscopic PDE)

\subsubsection{Parameterization}
A common technique used in numerical models describing complex physics phenomena is replacing dynamic systems that are hard to model with parameterized simple processes, which is named \textit{parameterization}. While the traditional way to decide parameters is to minimize the discrepancy between models' output and observed data by conducting grid search or Bayesian inference within a certain parameter space, recent works have adopted advances in machine learning as the approximation of unknown dynamics. 

In geology, \cite{chan2017parametrization} study the application of Wasserstein GAN~\cite{arjovsky2017wasserstein} for the parametrization of geological models. The effectiveness of the method is assessed for uncertainty propagation tasks using several test cases involving different permeability patterns and subsurface flow problems.
\cite{goldstein2014data} develop a new predictor for near-bed suspended sediment reference concentration under unbroken waves using genetic programming, a machine learning technique.

In meteorological science, \cite{brenowitz2018prognostic} show that a neural network‐based parameterization is successfully trained using a near‐global aqua‐planet simulation with a 4‐km resolution (NG‐Aqua).
The neural network predicts the apparent sources of heat and moisture averaged onto (160 km$^2$) grid boxes.
\cite{gentine2018could} present a novel approach to convective parameterization based on machine learning, using an aquaplanet with prescribed sea surface temperatures as a proof of concept.
A deep neural network is trained with a super-parameterized version of a climate model in which convection is resolved by thousands of embedded 2‐D cloud resolving models.  

In chemistry, supervised learning on molecules has incredible potential to be useful in chemistry, drug discovery, and materials science. \cite{gilmer2017neural} reformulate existing models into a single common framework we call Message Passing Neural Networks and explore additional novel variations within this framework.
Using MPNNs we demonstrate state of the art results on an important molecular property prediction benchmark.
\cite{wang2019molecule} utilize molecular graph data for property prediction based on spatial graph convolution neural networks.

\subsubsection{Reduced-Order Models}

Instead of directly modeling the complex dynamics in the observation space, researchers have developed simplified but more interpretable models (reduced-order models, ROMs) in the hidden space, which is usually derived from the observation space via dimensionality reduction. In recent works, machine learning models have emerged as effective tools to discover such hidden spaces and more information preserving transformations between them and the observation space.

In dynamic system analysis, Koopman theory is a typical ROM. Koopman theory is based on the insight that the state space of a non-linear dynamic system can be encoded into an infinite-dimensional space where the dynamics is linear~\cite{koopman1931hamiltonian}. In practice, people assume the infinite-dimensional space can be approximated with a finite-dimensional space. The key problem is then to find a proper pair of encoder/decoder to map from/to the state space to/from the hidden space.
Traditionally, people construct the encoder/decoder with hand-crafted functions, such as the identity function in Dynamic Mode Decomposition (DMD)~\cite{schmid2010dynamic}, nonlinear functions in Extended DMD (EDMD)~\cite{williams2015data}, and kernel functions in Kernel DMD (KDMD)~\cite{kevrekidis2016kernel}. However, hand-crafted functions may fail to fit complex dynamic systems and are hard to design without domain-specific knowledge. Thus, recent works ~\cite{li2020learning,azencot2020forecasting,lusch2018deep} construct encoders/decoders using neural networks as trainable universal approximators. They demonstrate that the combination of neural networks and Koopman theory achieves comparable or even higher performance than the Koopman approximators with hand-crafted mapping functions, while enjoying the ability to generalize to multiple datasets with the same design.~\cite{li2020learning} further shows that the integration of Koopman theory allows the model to adapt to new systems with unknown dynamics faster than pure neural networks.

In fluid dynamics, ROMs is largely considered due to the unprecedented physical insight into turbulence offered by high-fidelity computational fluid dynamics (CFD).
\cite{xiao2019reduced} develop a dimensionality reduction method called Non-Intrusive Reduced Order Model (NIROM) for predicting the turbulent air flows found within an urban environment. 
\cite{mohan2018deep} demonstrate a deep learning based approach to build a ROM using the POD basis of canonical DNS datasets, for turbulent flow control applications.
They find that a type of Recurrent Neural Network, the Long Short Term Memory (LSTM) which has been primarily utilized for problems like speech modeling and language translation, shows attractive potential in modeling temporal dynamics of turbulence.

\subsubsection{Causality}
\paragraph*{Causal Discovery and Causal Inference in Time Series}
How to discover the underlying causal structure is a fundamental problem and has still been actively studied.
\cite{rubin1974estimating,pearl2009causality,imbens2015causal} introduce the problem and provided a mathematical framework for causal reasoning and inference under causal graphical models (also known as Bayesian networks (BN)~\cite{koller2009probabilistic}). 
\cite{granger1969investigating} formalize a concept of quantifiable causality in time series, called Granger causality.
From the pioneering works, learning causal associations from time series has been an emerging topic in machine learning and deep learning community as well.
\cite{runge2018causal} propose a method to distinguish direct from indirect dependencies and common drivers among multiple time series to reconstruct a causal network.
\cite{runge2019detecting} quantify causal associations in nonlinear time series and \cite{runge2019inferring,nauta2019causal} provide promising applications of the causal discovery in time series.
\cite{pamfil2020dynotears} introduce a smooth acyclicity constraint to multivariate time series inspired by~\cite{zheng2018dags} that consider a causal discovery as a purely continuous optimization problem.
 
 For causal inference part, G-computation formula, g-estimation of structural nested mean models \cite{hernan2010causal}, and inverse probability of treatment weighting in marginal structural models (MSMs) \cite{robins2000marginal, fitzmaurice2008estimation} rely on linear predictors for estimation to estimate treatment effects. Recurrent marginal structural networks (RMSNs) \cite{lim2018forecasting} is proposed to further improve MSM's ability by capturing nonlinear dependencies. In addition, Gaussian process~\cite{schulam2017reliable,soleimani2017treatment} has been tailored to estimate treatment response in a continuous-time settings in order to incorporate non-deterministic quantification. Furthermore, \cite{pearl2012measurement} and \cite{kuroki2014measurement} theoretically prove that observed proxy variables can be used to capture hidden confounders and estimate treatment effects. TSD \cite{bica2020tsd} introduces recurrent neural networks in the factor model to estimate the dynamics of confounders. In a similar vein, \cite{hatt2021sequential} propose a sequential deconfounder to infer hidden confounders  by using Gaussian process latent variable model.  DTA~\cite{kuzmanovic2021deconfounding} estimates treatment effects under dynamic setting using observed data as noisy proxies. Besides, DSW \cite{liu2020estimating}  infers the hidden confounders by using a deep recursive weighted neural network that combines current treatment assignment and historical information. DNDC \cite{ma2021deconfounding} aims to learn how hidden confounders behave over time by using current network observation data and historical information.
Although many works are successful to discover unknown causal structure from observational data directly and make effective causal inference, there are few works to leverage explicit causal relations from physical knowledge to improve data-driven models. 

\paragraph*{Counterfactual Analysis of Physical Dynamics} 
Counterfactual analysis in the physical world is typically concerned with analytically predicting the effects of various types of interventions/treatments, including the physical laws of current environment~\cite{battaglia2016interaction, chang2016compositional, wu2015galileo}, the actions of the agent itself~\cite{levine2014learning, li2019multi, wahlstrom2015pixels, watter2015embed}, and the decision outcomes of other agents in multi-agent systems~\cite{he2016opponent, tian2019regularized}. With the development of 3D simulation in the field of machine learning, several benchmarks performing counterfactual analysis in physics have emerged. CLEVRER~\cite{Yi*2020CLEVRER:} asks agents to answer a counterfactual question after observing a video showing the movement and collision of a 3D object. Physics 101~\cite{phys101} presents a video benchmark containing over 101 real-world objects physically interacting in four different physical scenarios. \cite{li2020causal} proposes a counterfactual benchmark with two tasks: a scene where balls interact according to unknown interaction laws (e.g., gravity or elasticity), and a scene where clothes are folded by the wind. The agent needs to discover causal relationships between counterfactual variables and objects and then predicts future frames. CoPhy~\cite{Baradel2020CoPhy:} separates the observed experiments from those of counterfactuals and includes three complex 3D scenarios involving rigid body dynamics.

The counterfactual analysis of the physics-informed models relies on the separation between the physical information features and the remaining features~\cite{villegas17mcnet, NIPS2017_2d2ca7ee}  and may incorporate additional information based on the available prior knowledge of the scene  ~\cite{ villegas2017learning}. Recently, PhyDNet~\cite{guen2020disentangling} explicitly disentangles PDE dynamics from unknown complementary information. The interpretable intuitive physical model~\cite{ye2018interpretable} proposes an encoder-decoder framework for predicting future collision event frames. The encoder layer infers physical properties, such as mass and friction, from the input frames. Then, the decoder decomposes the potential physical vectors by outputting optical flow. PIP~\cite{duan2021pip} uses a deep generative model to build approximate mental simulations by generating a framework for future physical interactions and then employs selective temporal attention in the form of spanwise selection to predict the outcome of physical interactions. CWMs ~\cite{li2020causal} allow unsupervised modeling of relationships between the intervened observations and the alternative futures by learning an estimator of the latent confounding factors. Cophy ~\cite{Baradel2020CoPhy:} predicts alternative outcomes of a physical experiment by estimating the potential performance of confounding factors. Filter-Cophy ~\cite{janny2022filteredcophy} further learns and acts on a suitable hybrid potential representation based on a combination of dense features, sets of 2D keypoints, and an additional latent vector per keypoint.


\subsection{Physics for ML: Improvement of Data-Driven Models from External Knowledge}
Data-driven methods represented by machine learning and neural networks have achieved great success recently in a wide range of real-world problems due to the universal approximation ability of neural networks~\cite{hornik1989multilayer}, the increase of available data for model training, and the rapid development of hardware neural network accelerators. However, the optimization of neural networks is highly non-convex, and its convergence to global minima is hard to achieve in practice. Optimization processes converging to local minima without constraints may lead to models with limited generalization ability or results violating existing knowledge including commonsense, logic rules, and physics laws~\cite{von2019informed}. In this work, we focus on the integration of physics knowledge in machine learning, and the following sections will introduce the integration in multiple domains of machine learning respectively.

\subsubsection{Supervised Learning}
Typical supervised learning tasks including classification and forecasting cover various real-world applications from multi-agent systems, computer vision, time series analysis, and spatio-temporal data modeling. While deep neural network-based methods have been the dominating solution in these areas recently, people have also explored several novel approaches with the incorporation of physics knowledge in tasks on data from physics-related scenarios, such as object-centric data, spatio-temporal data, and geometry data.

\paragraph{Object-centric Data}
Object-centric data is generated by systems composed of multiple discrete objects. Examples of real-world object-centric data include trajectory data from multi-agent systems~\cite{greydanus2019hamiltonian, kipf2018neural,li2020generative}, position and velocity data from networks of motion sensors~\cite{kipf2018neural,yan2018spatial}, and molecule data~\cite{gilmer2017neural}. Graph neural network (GNN)-based methods~\cite{battaglia2016interaction,battaglia2018relational,sanchez2018graph,li2020generative} achieve state-of-the-art results on most tasks with object-centric data due to the matching of its inductive bias~\cite{battaglia2018relational} and the interacting physics property in object-centric data.~\cite{xu2020can} demonstrates that GNNs generalize well in many popular object-centric tasks because the forwarding process of GNNs aligns to the underlying reasoning process. In addition to this alignment, other types of physics knowledge are also integrated in recent works.~\cite{schutt2017schnet} obtains a joint model for the total energy and interatomic forces of molecules that follows the energy-conserving law.~\cite{lutter2018deep} proposes Deep Lagrangian Networks (DeLaN) for robotics, on which Lagrangian Mechanics is imposed and physical plausibility is maintained.~\cite{greydanus2019hamiltonian} designs models for forecasting n-body systems that learn and respect exact conservation laws - Hamiltonian mechanics - in an unsupervised manner. Lagrangian Neural Networks (LNNs) in~\cite{cranmer2020lagrangian} further parameterize arbitrary Lagrangians using neural networks without canonical coordinates or restrictions of the functional form of learned energies, and can be applied to graphs and continuous systems.~\cite{li2020learning} learns compositional Koopman operators using GNNs and shows better efficiency and generalization than existing GNN baselines in multi-object systems such as ropes and soft robots.

\paragraph{Spatio-Temporal Data}
Spatio-temporal data records the dynamics of values of interest in multiple locations within a period of time. The most common form of spatio-temporal data is video, where all locations are aligned to rectangular mesh grids in some specific area. Spatio-temporal data can also be generated from networks of irregularly spaced sensors in domains such as traffic~\cite{li2018dcrnn_traffic}, weather~\cite{zhou2021informer}, and electricity~\cite{wu2020connecting}. Since the underlying processes of spatio-temporal data are usually governed by physics laws, physics-informed methods have potentials to further improve the performance of neural network models.~\cite{guen2020disentangling} explicitly disentangles known PDE dynamics from unknown factors and performs PDE-constrained prediction in latent-space, both of which contributes to video forecasting performance.~\cite{He2020AdvectiveNet:} mimics the pipeline of physical flow simulation, and evolves and accumulates point features in point clouds using flow velocities generated from a high-dimensional force field. It demonstrates the efficacy of the proposed method in various point cloud classification and segmentation problems.~\cite{Seo*2020Physics-aware} learns finite differences of sparsely available data points inspired by physics equations, and shows the superiority in synthetic graph signal prediction and real-world weather forecasting tasks.~\cite{iakovlev2021learning} proposes a general continuous-time differential model for dynamical systems, which admits arbitrary space and time discretizations, and enables efficient neural PDE inference. 

\paragraph{Manifold Data}
Manifold data describes signals defined on non-planar surfaces such as spheres and surfaces of complex 3D objects, where the Euclidean geometry only holds locally near each point. Examples are magnetoencephalography (MEG) brain activity signal data~\cite{Defferrard2020DeepSphere}~(on sphere), omnidirectional image data~\cite{armeni2017joint}~(on 3D objects), and human scan data~\cite{bogo2014faust}~(on 3D objects). Without the restriction that data points are distributed on planes, general manifold data comes with richer information about the space including local structures and symmetries of the space. Meanwhile, existing convolution-based neural network architectures are either no longer applicable in non-planar manifolds (CNNs) or not capable of fully exploiting the spatial information (GNNs). Instead, physics-informed neural networks incorporate established knowledge of manifolds into the construction of new types of convolutions for manifolds and bridge the gap. \cite{masci2015geodesic}~constructs intrinsic CNN-like architectures on non-planar surfaces under the geodesic polar coordinate system. \cite{monti2017geometric}~alternatively uses the principal curvature directions as fixed gauges to construct convolutions in corresponding tangent fields. \cite{boscaini2016anisotropic}~constructs convolution kernels as rotating filters and collects the strongest responses among all possible directions. \cite{cohen2019gauge,de2020gauge}~further extends convolution filters to be gauge-equivariant via parallel transporting geometric features to the same vector space before applying the filters.

\subsubsection{Model-Based Control}
Due to the integration of physics knowledge, physics-informed machine learning models enjoy better physical plausibility, higher data efficiency and stronger generalization ability compared to pure neural network models, all of which are critical properties for constructing a good model describing the relation between the control input and the state transition of dynamic systems. Recent works have shown that physics-informed machine learning models achieve significant success in model predictive control and model-based reinforcement learning. \cite{shi2019neural}~integrates a deep neural network (DNN) that learn high-order interactions into the dynamics model, and constrain the Lipschitz constant of the DNN to guarantee system stability. \cite{lutter2018deep,zhong2019symplectic}~enforce Lagrangian and Hamiltonian dynamics in the modeling of underlying system dynamics respectively, and both outperform model learning approaches without physics knowledge in trajectory tracking error, learning speed, and robustness. \cite{li2020learning}~shows that dynamics learned via compositional Koopman operators can quickly adapt to new environments of unknown physical parameters in online learning. \cite{holl2019learning,yin2021augmenting}~augment partial differential equations (PDEs) that approximately describe continuous physical systems with controllable force terms, and demonstrate that proposed methods successfully control the evolution of complex physical systems.

\section{Physics Knowledge in PIML}
\label{sec:physics-knowledge-in-piml}
In this section, we introduce several categories of general physics knowledge integrated in PIML. While there is much more domain/task-specific knowledge that can be incorporated for corresponding solutions, each category we introduce in this section covers a wide range of problems and inspires a series of works generally applicable to them instead of leading to only one or two task-specific solutions.


\subsection{Classical Mechanics and Energy Conservation Laws}
\label{sec:classical-mechanics-and-energy-conservation-laws}
Newtonian, Lagrangian, and Hamiltonian mechanics are three typical approaches describing systems of classical mechanics. While the Newtonian mechanics has been widely used to describe the relations among locations, velocities, accelerations and forces, Lagrangian and Hamiltonian mechanics provide effective tools to enforce laws of conservation of energy in the modeling of dynamic systems. Since Newtonian mechanics described as the famous Newton's Three Laws has been widely known, here we only focus on Lagrangian and Hamiltonian mechanics.

\paragraph{Lagrangian Mechanics}
The Lagrangian mechanics defines the Lagrangian function $L$ of generalized coordinates $\vq$ and its gradient w.r.t time $\dot{\vq}$ to fully describe dynamics of a mechanical system. Usually $L$ is chose as the difference between the kinetic energy $T$ and the potential energy $V$, i.e. $L(\vq,\dot{\vq})\equiv T(\vq, \dot{\vq}) - V(\vq)$.

The defined Lagrangian function must satisfy the principle of stationary action: the real physical trajectory of a system will always take is the one in which the action happens to be stationary, which is mathematically expressed as:
\begin{equation}
\label{eq:stationary-action}
    \delta\int_{t_1}^{t_2}Ldt= 0.
\end{equation}
We can further derive the important Euler-Lagrange equation from Eq~\ref{eq:stationary-action}:
\begin{equation}
    \label{eq:euler-lagrange}
    \frac{d}{dt}\nabla_{\dot{\vq}}L = \nabla_{\vq}L.
\end{equation}

According to Noether's theorem, the energy of a system is conserved if the system has a time-translation symmetry, i.e. if the Lagrangian function does not explicitly depend on time ($\frac{\partial L}{\partial t} = 0$). Therefore, as long as we model the Lagrangian function as a function of $\vq, \dot{\vq}$ without the explicit time term $t$ and satisfies Eq~\ref{eq:euler-lagrange}, it automatically satisfies the law of conservation of energy. With Eq~\ref{eq:euler-lagrange} we can derive the expression of $\ddot{\vq}$ with the chain rule~\cite{cranmer2020lagrangian}:
\begin{equation}
    (\nabla_{\dot{\vq}}\nabla_{\dot{\vq}}^TL)\ddot{\vq} + (\nabla_{{\vq}}\nabla_{\dot{\vq}}^TL)\dot{\vq} = \nabla_{\vq}L.
\end{equation}
When $L$ is modeled as a differential function such as neural networks, we can solve $\ddot{\vq}$ as:
\begin{equation}
    \ddot{\vq} = (\nabla_{\dot{\vq}}\nabla_{\dot{\vq}}^TL)^{-1}[\nabla_{\vq}L - (\nabla_{{\vq}}\nabla_{\dot{\vq}}^TL)\dot{\vq}].
\end{equation}
Usually one system's $\vq$ and $\dot{\vq}$ can be observed as spatial coordinates and speeds of objects, thus the system can be solved with $(\vq, \dot{\vq})\vert_{t=T} = \int_{0}^{T}(\dot{\vq}, \ddot{\vq})dt$.

For systems with non-conservative forces, Eq~\ref{eq:euler-lagrange} can be extended as
\begin{equation}
    \frac{d}{dt}\nabla_{\dot{\vq}}L - \nabla_{\vq}L = \bm{\tau},
\end{equation}
where $\bm{\tau}$ are generalized forces and can be used for controllers such as a PD-Controller~\cite{lutter2018deep}.

\paragraph{Hamiltonian Mechanics}
The Hamiltonian mechanics defines the Hamiltonian function $H$ as the function of a pair of variables $(\vq, \vp)$, where $\vq$ is the vector of spatial coordinates of system objects and $\vp$ is the vector of their momentum. $(\vq, \vp)$ must satisfy the canonical condition:
\begin{equation}
    \label{eq:canonical-cond}
    \vp = \nabla_{\dot{\vq}}L,
\end{equation}
where $L$ is the Lagrangian function of the system.
The Hamiltonian function is defined as follows:
\begin{equation}
    \label{eq:hamiltonian}
    H(\vq, \vp) = \dot{\vq}\cdot\vp - L.
\end{equation}
Combine Eq~\ref{eq:canonical-cond}, Eq~\ref{eq:hamiltonian} and Eq~\ref{eq:euler-lagrange} we can derive:
\begin{equation}
\label{eq:hamiltonian-derive}
    \dot{\vp} = -\nabla_{\vq}H,~\dot{\vq} = \nabla_{\vp}H.
\end{equation}
With Eq~\ref{eq:hamiltonian-derive}, we can verify that
\begin{equation}
\label{eq:hamiltonian-conservative}
    \frac{dH}{dt} = \dot{\vq}\nabla_{\vq}H + \dot{\vp}\nabla_{\vp}H = 0.
\end{equation}
In the context of classical mechanics, we have $\vp = M\dot{\vq}$, $T=\frac{1}{2}\dot{\vq}^TM\dot{\vq}$,  $L=T-V$, then Eq~\ref{eq:hamiltonian} can be rewritten as $H(\vq, \vp) = \dot{\vq}^TM\dot{\vq} - T + V = T + V$, 
where $M$ is the mass matrix of the system. Here the Hamiltonian function $H$ is exactly the total energy of the system. Therefore, Eq~\ref{eq:hamiltonian-conservative} shows that systems described with Hamiltonian mechanics follows the law of conservation of energy.

\subsection{Symmetry, Invariant and Equivariant Functions}
\label{sec:symmetry}
A symmetry defined on an object or system is some transformation that keeps certain properties unchanged. Typical symmetries include shifts in visual object classification problems, rotations in molecule property prediction problems, and permutations in particle systems. For one object or system, its symmetries form a \textit{symmetry group}, where the following rules are satisfied: (1) associativity, (2) identity, (3) inverse, (4) closure. Some common symmetry groups are: $T(n)$ ($n$-dim translation group), $O(n)$ ($n$-dim distance-perserving group including both rotation and reflection), $SO(n)$ ($n$-dim rotation group), $E(n)$ ($n$-dim Euclidean group, including translation, rotation and reflection).

On a set $\Omega$ representing a domain, we can define a \textit{group action} as a mapping $(\mathfrak{g}, u)\mapsto \mathfrak{g}.u$, where $\mathfrak{g}$ is one element of a symmetry group $\mathfrak{G}$, $u$ and $\mathfrak{g}.u$ are two points in $\Omega$. A group action shall satisfy associativity: $\mathfrak{g}.(\mathfrak{h.u}) = (\mathfrak{gh}).u$ for all $\mathfrak{g,h}\in\mathfrak{G}$ and $u\in\Omega$.
In comparison, \textit{group action} on a signal space $\mathcal{X}:\Omega\rightarrow\mathbb{R}^n$ is defined as:
\begin{equation}
    (\mathfrak{g}.x)(u) = x(\mathfrak{g}^{-1}u).
\end{equation}
We can verify that the above definition satisfies the associativity:
\begin{equation}
    (\mathfrak{g}.(\mathfrak{h}.x))(u) = (\mathfrak{h}.x)(\mathfrak{g}^{-1}(u)) = x((\mathfrak{gh})^{-1}u) = ((\mathfrak{gh}).x)(u).
\end{equation}

On a domain $\Omega\in\mathbb{R}^n$, we can define an $n$-dimensional real \textit{representation} of a group $\mathfrak{G}$ as a map $\rho:\mathfrak{G}\rightarrow\mathbb{R}^{n\times n}$, connecting each $\mathfrak{g}\in\mathfrak{G}$ to an invertible matrix $\rho(\mathfrak{g})$, and satisfying the associtivity $\rho(\mathfrak{gh})=\rho(\mathfrak{g})\rho(\mathfrak{h})$ for all $\mathfrak{g,h}\in\mathfrak{G}$.

For a symmetry group $\mathfrak{G}$ we have the definitions of $\mathfrak{G}$-invariant and $\mathfrak{G}$-equivariant. A function $f:\mathcal{X}(\Omega)\rightarrow\mathcal{Y}$ is $\mathfrak{G}$-invariant if $f(\rho(\mathfrak{g})x) = f(x)$ for all $\mathfrak{g}\in\mathfrak{G}$ and $x\in\mathcal{X}(\Omega)$, and it is $\mathfrak{G}$-equivariant if $f(\rho(\mathfrak{g})x)=\rho(\mathfrak{g})f(x)$.

By stacking multiple neural network layers(functions), each of which satisfies either equivariance or invariance under some symmetry group, we can incorporate the knowledge of symmetries of domains into the resulting network. Table~\ref{tab:geodl}~\cite{bronstein2021geometric} connects some neural network architectures with their corresponding symmetry groups and domains.

\begin{table}[htbp]
    \centering
    \begin{tabular}{lll}
    \toprule
    Architecture & Symmetry group $\mathfrak{G}$ & Domain $\Omega$\\
    \midrule
    CNN & Grid & Translation\\
    Spherical CNN~\cite{s.2018spherical} & Sphere / SO(3) & Rotation SO(3)\\
    Intrinsic Mesh CNN~\cite{masci2015geodesic,monti2017geometric} & Manifold & Isometry Iso($\Omega$) / Gauge symmetry SO(2)\\
    GNN & Graph & Permutation $\Sigma_n$\\
    Deep Sets~\cite{zaheer2017deep} & Set & Permutation $\Sigma_n$\\
    Transformer & Complete Graph & Permutation $\Sigma_n$\\
    LSTM & 1D Grid & Time warping\\
    \bottomrule
    \end{tabular}
    \vspace{1ex}
    \caption{Neural network architectures, symmetry groups and domains.}
    \label{tab:geodl}
\end{table}

\subsection{Numerical Methods for Partial Differential Equations (PDEs)}
\label{sec:numerical-pde}
A generic form of PDEs describing the evolution of a continuous value $v(x,t)$ is as follows:
\begin{equation}
    \label{eq:generic-pde}
    \frac{\partial v}{\partial t} = F(t,x,v,\frac{\partial v}{\partial x_i}, \frac{\partial^2v}{\partial x_i \partial x_j}, \dots).
\end{equation}
\paragraph{Finite Difference Method}
Finite difference methods approximate spatial derivatives on the right hand side of Eq~\ref{eq:generic-pde} at a certain point $x_0$ as a linear combination of function values at $N$ neighbors of $x_0$. In 1D case, the approximation of the $l$-th order spatial derivative can be formulated as (here we omit $t$):
\begin{equation}
    \left.\frac{\partial^l f}{\partial x^l}\right\vert_{x=x_0} = \sum_{n=1}^N \alpha_nf(x_0 + h_n).
\end{equation}
$\alpha_1, \alpha_2, \dots, \alpha_n$ can be solved via expanding $\alpha_1f(x_0+h_1), \alpha_2f(x_0+h_2), \dots, \alpha_Nf(x_0+h_N)$ at $x_0$ to the $N$-th order:
\begin{equation}
    \left\{
    \begin{array}{cc}
        \alpha_1f(x_0 + h_1) &= \alpha_1\sum_{k=0}^{N-1}\frac{1}{k!}\frac{\partial^kf}{\partial x^k}h_1^k + O(h_1^{N-1})  \\
        \dots\\
        \alpha_Nf(x_0 + h_N) &= \alpha_N\sum_{k=0}^{N-1}\frac{1}{k!}\frac{\partial^kf}{\partial x^k}h_N^k + O(h_N^{N-1})  \\
    \end{array}
    \right.
    ,
\end{equation}
which can be summed together as 
\begin{equation}
    \sum_{n=1}^N \alpha_nf(x_0 + h_n) = \sum_{k=0}^{N-1}(\frac{1}{k!}\sum_{i=1}^N\alpha_ih_i^k)\frac{\partial^kf}{\partial x^k} + O(\max_{i\in[N]}h_i^{N-1})
\end{equation}
To solve $\alpha_1, \alpha_2, \dots, \alpha_n$ for a certain order $l$, we can let the multiplier of $\frac{\partial^k f}{\partial x^k}$ be 1 and others be 0, and solve following linear equations:
\begin{equation}
    \left[
    \begin{array}{cccc}
         1 & 1 & \dots & 1\\
         h_1^1 & h_2^1 & \dots & h_N^1\\
         \vdots & \vdots & & \vdots\\
         h_1^l & h_2^l & \dots & h_N^l\\
         \vdots & \vdots & & \vdots\\
         h_1^{N-1} & h_2^{N-1} & \dots & h_N^{N-1}
    \end{array}
    \right]
    \left[
    \begin{array}{c}
         \alpha_1\\
         \alpha_2\\
         \vdots\\
         \alpha_N
    \end{array}
    \right] =
    \left[
    \begin{array}{c}
         0!\cdot 0\\
         1!\cdot 0\\
         \vdots\\
         l!\cdot 1\\
         \vdots\\
         (N-1)!\cdot 0
    \end{array}
    \right],
    \label{eq:coef-linear}
\end{equation}
the solution $\alpha_1^*, \dots, \alpha_N^*$ of which satisfies
\begin{equation}
    \frac{\partial^kf}{\partial x^k} = \sum_{n=1}^N \alpha_n^*f(x_0 + h_n) + O(\max_{i\in[N]}h_i^{N-1}).
\end{equation}
The above method can be extended to multi-dimensional cases with multi-dimensional Taylor expansion. After approximating spatial derivatives on the left hand side, Eq~\ref{eq:generic-pde} can be solved with standard techniques of numerical integration. While Eq~\ref{eq:coef-linear} can fully determine all coefficients, recent works~\cite{long2018pde,long2019pde} relax it by removing some constraining equations and use neural networks to learn undetermined coefficients to combine prior knowledge with stronger expressivity.

\paragraph{Finite Volume Method}
Unlike finite difference method, finite volume method represent the value $v(x,t)$ with its averages over a grid cell: $v_i(t) = \Delta x^{-1}\int_{x_i-\Delta x/2}^{x_i + \Delta x/2}v(x^\prime, t)dx^\prime$. Similar to the finite difference method, spatial derivatives can also be estimated as the linear combination of averaged cell values: $\frac{\partial^l v}{\partial x^l}=\sum_{i=1}^N\alpha_i^{(l)}v_i$, and the coefficients are estimated in the same way as finite difference method.

For finite volume methods, the equation must be able to be rewritten as:
\begin{equation}
    \frac{\partial v}{\partial t} = \frac{\partial}{\partial x}J(t,x,v,\frac{\partial v}{\partial x_i}, \frac{\partial^2v}{\partial x_i \partial x_j}, \dots),
\end{equation} where J is called a \textit{flux} and has an analytical form derived from the original PDE. The evolution along time can be carried out in following steps: (1) spatial derivatives are estimated on the boundary between grid cells; (2) the flux J is calculated with approximated derivatives using its analytical form; (3) the temporal derivative of averaged cell values is calculated via subtracting J ath the cell's left and right boundaries. The final step can be conducted with techniques that promote stability, such as monotone numerical fluxes and Godunov flux~\cite{bar2019learning}. Recent works have integrated data-driven models in step (1) to improve the estimation of spatial derivatives. For example,~\cite{bar2019learning} estimate coefficients with the combination of neural network results and numerical results, and~\cite{wang2019learning} uses policies trained via reinforcement learning to determine coefficients for estimation.

\paragraph{Finite Element Method}
Finite element method divide a space into small parts (elements) and approximate the PDE on each. As illustrated in~\cite{xue2020amortized}, we use the Poisson's equation to show the basic idea of finite element method.

The formulation of the Poisson's equation is as follows:
\begin{equation}
\begin{aligned}
     -\Delta u &= \lambda\quad\mathrm{in}~\Omega\\
     u &= 0\quad\mathrm{on}~\Gamma.
\end{aligned}
\end{equation}

In finite element method, we multiply a \textit{test function} $v$ on both sides, integrate over $\Omega$, and use integration in parts to derive the \textit{weak formulation}:
\begin{align}
\label{eq:weak-formulation}
    a(u,v) &= l(\lambda, v),~\mathrm{for~all}~v\in\mathbb{V},\\
    \mathrm{where}~a(u,v) &= \int_{\Omega}\nabla u\cdot \nabla v dx,~l(\lambda, v) = \int_{\Omega}\lambda v dx.
\end{align}
Then we construct a finite-dimensional subspace $\mathbb{V}_h \subset \mathbb{V}$, where $\mathbb{V}_h$ is a piece-wise polynomial function space spanned via $\left\{ \phi_1,\phi_2,\dots,\phi_n \right\}$. To solve the weak formulation, we transform it to the \textit{Galerkin weak formulation}, which is an approximation of~Eq \ref{eq:weak-formulation}:
\begin{equation}
\label{eq:galerkin-weak-form}
    \mA\vu=\vf,
\end{equation}
where $\mA\in\mathbb{R}^{n\times n}$, $A_{ij} = a(\phi_i, \phi_j)$, $\vu\in\mathbb{R}^n$ is the solution vector and $\vf\in\mathbb{R}^n$ is the source vector with $f_i=l(\phi_i, \lambda)$. The solution of Eq~\ref{eq:galerkin-weak-form} gives the best solution of the PDE in $\mathbb{V}_h$.

\subsection{Koopman Theory}
Given a non-linear dynamic system with its state vector at time $t$ denoted as $\vs_t\in\sR^m$. The system can be described as $\vs_{t+1} = F(\vs_t)$. As defined in~\cite{koopman1931hamiltonian}, the Koopman operator $\mathcal{K}_F$ is a linear transformation defined on a function space $\mathcal{F}$ by $\mathcal{K}_Fg = g\circ F$ for every $g: \sR^m \rightarrow \sR$ that belongs to the infinite-dimensional Hilbert space $\mathcal{F}$. With the definition we have $\mathcal{K}_Fg(\vs_t) = g\circ F(\vs_t) = g(\vs_{t+1})$.

The Koopman theory~\cite{koopman1931hamiltonian} guarantees the existence of $\mathcal{K}$, but in practice we often assume the existence of an invariant finite-dimensional subspace $\mathcal{G}$ of $\mathcal{F}$ spanned by $k$ bases $\{g_1, g_2,\dots, g_k\}$. Define $\vg_{t} = [g_1(\vs_t), g_2(\vs_t), \dots, g_k(\vs_t)]^T$ and $\vg_{t+1} = [g_1(\vs_{t+1}), g_2(\vs_{t+1}), \dots, g_k(\vs_{t+1})]^T$, under the assumption we have $\vg_t, \vg_{t+1}\in \mathcal{G}$ and there exists a Koopman matrix $\mK\in\sR^{k\times k}$ s.t. $\vg_{t+1} = \mK\vg_t$. The key problem is to find the pair of mappings between the state space $\sR^m$ and the invariant subspace $\mathcal{G}\in\sR^k$: $\vg: \sR^m \rightarrow \sR^k$ and $\vg^{-1}: \sR^k \rightarrow \sR^m$. Recent works~\cite{lusch2018deep,li2020learning,azencot2020forecasting} utilize neural networks as $\vg$ and $\vg^{-1}$ to find the mappings in a data-driven way.
\section{Methods of PIML}
\label{sec:methods-of-piml}
Typical solutions to a problem with machine learning involve three key parts: data, model, and optimization, each of which can be integrated with prior physics knowledge. In the following parts, we will introduce existing techniques of incorporating physics knowledge to each part respectively. However, we should notice that these techniques are not mutually exclusive: physics knowledge can be integrated to more that one parts of the machine learning solution.

\begin{table}[htbp]
    \centering
    \resizebox{\linewidth}{!}{
    \begin{tabular}{@{}|c|c|c|c|c|c|c|c|@{}}
        \hline
        \multirow{2}{*}{\diagbox{Knowledge Form}{Integration Method}} & \multicolumn{3}{c|}{Data} & \multicolumn{2}{c|}{Model} & \multicolumn{2}{c|}{Optimization} \\ \cline{2-8}
         & Simulation Data & Transfer Learning & Auxiliary Tasks & Computation Graph & Fusion & \begin{tabular}[c]{@{}c@{}}Knowledge\\Based Loss\end{tabular} & Regularization \\ \hline
         \begin{tabular}[c]{@{}c@{}}Domain Knowledge\\in Analytical Form\end{tabular} & \begin{tabular}[c]{@{}c@{}}\cite{rasp2020weatherbench}\cite{li2018learning}\cite{sanchez2020learning}\cite{pfaff2020learning}\\\cite{gronquist2021deep}\cite{kashinath2021physics}\cite{sanchez2020learning}\cite{li2018learning}\\\cite{ummenhofer2019lagrangian}\cite{pfaff2021learning}\cite{belbute2020combining}\cite{sharifi2019downscaling}\\\cite{vandal2017deepsd}\cite{lee2020coarse}\cite{mohan2018deep}\end{tabular} & \cite{tremblay2018training}\cite{bousmalis2018using}\cite{mueller2018driving}\cite{jia2021physics} & \cite{schutt2017schnet} & \cite{wang2020towards}\cite{jia2021physics}\cite{He2020AdvectiveNet:} & \cite{garcia2019combining}\cite{belbute2020combining} & \begin{tabular}[c]{@{}c@{}}\cite{raissi2017physicsI}\cite{raissi2017physicsII}\cite{raissi2018deep}\cite{raissi2019physics}\\\cite{esmaeilzadeh2020meshfreeflownet}\cite{Shi_Mo_Di_2021}\cite{jia2021physics}\cite{yang2019enforcing}\cite{wu2020enforcing}\end{tabular} & \cite{shi2019neural}\cite{holl2019learning}\cite{yin2021augmenting} \\ \hline
         \begin{tabular}[c]{@{}c@{}}Energy Conservation\\Law\end{tabular} & \notableentry & \notableentry & \notableentry & \cite{lutter2018deep}\cite{cranmer2020lagrangian}\cite{greydanus2019hamiltonian}\cite{Toth2020Hamiltonian}\cite{zhong2019symplectic} & \notableentry & \notableentry & \notableentry \\ \hline
         Symmetry & \notableentry & \notableentry & \notableentry & \begin{tabular}[c]{@{}c@{}}\cite{satorras2021n}\cite{schutt2017schnet}\cite{horie2021isometric}\cite{weiler2019general}\\\cite{thomas2018tensor}\cite{fuchs2020se}\cite{finzi2020generalizing}\cite{hutchinson2021lietransformer}\\\cite{weyn2020improving}\cite{masci2015geodesic}\cite{monti2017geometric}\cite{boscaini2016anisotropic}\\\cite{cohen2019gauge}\cite{de2020gauge}\end{tabular} & \notableentry & \notableentry & \notableentry \\ \hline
         \begin{tabular}[c]{@{}c@{}}Numerical Methods\\for PDEs\end{tabular} & \notableentry & \notableentry & \cite{thanasutives2021adversarial}\cite{seo2020physics} & \begin{tabular}[c]{@{}c@{}}\cite{long2018pde}\cite{long2019pde}\cite{jiang2018spherical}\cite{Seo*2020Physics-aware}\\\cite{bar2019learning}\cite{wang2019learning}\cite{alet2019graph}\cite{xue2020amortized}\\\cite{trask2019gmls}\cite{kashinath2020enforcing}\cite{um2020sol}\end{tabular} & \cite{long2018hybridnet}\cite{long2018pde}\cite{guen2020disentangling}\cite{Seo*2020Physics-aware}\cite{iakovlev2021learning} & \notableentry & \notableentry \\ \hline
         Koopman Theory & \notableentry & \notableentry & \notableentry & \cite{lusch2018deep}\cite{takeishi2017learning}\cite{li2020learning}\cite{azencot2020forecasting} & \notableentry & \notableentry & \notableentry \\
         \hline
    \end{tabular}}
    \vspace{1ex}
    \caption{Existing works classified based on forms of physics knowledge and integration methods. For certain types of knowledge forms, inductive bias integrated in reusable computation graphs have advantages over integration with data and optimization. }
    \label{tab:existing-works-kw-intm}
\end{table}

In Table~\ref{tab:existing-works-kw-intm}, we classify existing works based on the forms of physics knowledge and the integration methods. We notice that for domain knowledge taking analytical forms, existing works integrate the knowledge into all three aspects including data, model, and optimization. However, research works on integrating other general types of physics knowledge, including energy conservation law, symmetry, numerical methods for PDEs, and Koopman theory, mainly focus on incorporating corresponding knowledge into computation graphs. The main reason is that such general physics knowledge is possible to be transformed to inductive bias in reusable network architectures, which has advantages over data augmentation and physics knowledge based loss functions in terms of prediction performance and data efficiency~\cite{bronstein2021geometric}. This is due to that (1) general physics knowledge applies to various problems and thus leads to general network architectures, and (2) has simpler forms that can be translated to combinations of a limited number of differentiable operators compared to complex numerical simulators designed for domain-specific problems such as weather and turbulence.





\subsection{Physics-Informed Data Enhancement}
\subsubsection{Data Generated from Simulation}
The Universal Approximation Theorem~\cite{hornik1989multilayer} guarantees that multilayer neural networks with as few as one hidden layer can approximate any continuous function from one finite dimensional space to another to any desired degree of accuracy. Therefore, one straight forward method to incorporate physics knowledge into neural networks is to generate training data from the desired physics knowledge. When the data amount is abundant, the neural network is expressive enough and trained properly, the trained neural network will be able to approximate the behavior of the physics knowledge governing the data generation. Usually the neural network models can be accelerated with hardware such as GPU, FPGA and ASIC, thus they can act as good surrogate models with much lower computation costs while maintaining comparable accuracy to the numerical simulation.

\cite{rasp2020weatherbench} present a benchmark dataset from results of numerical global weather simulation with high computation cost, and provide scores of deep learning models. Results show that data-driven models trained with simulation data can achieve competitive results compared to numerical solutions while enjoying lower computation costs. \cite{li2018learning,sanchez2020learning,pfaff2020learning} utilize different variants of message passing graph neural networks~\cite{li2019propagation,battaglia2018relational,sanchez2018graph} respectively and train them with the simulation data of particle systems. Compared to the simulation methods generating datasets, the trained surrogate model can accurately predict the dynamics of a wide range of physical systems within the same architecture, and runs orders of magnitude faster.

\subsubsection{Transfer Learning}
For real-world tasks suffering from data limitation or labeling difficulties, the integration of prior physics knowledge about the tasks is critical. Simulators constructed with such physics knowledge can provide large amount of data with high label quality, and can be used to pre-train models. However, the differences between the target real-world data distributions and simulation data distributions call for techniques of transfer learning to mitigate the gap.

\cite{tremblay2018training} present a system for training object detection models with synthetic images. This work adopts the technique of domain randomization, where important parameters of simulators - including lighting, pose, object textures - are randomized in non-realistic ways to encourage the model to learn essential features.~\cite{bousmalis2018using} uses off-the-shelf simulators to render synthetic data for training a grasping system together with pixel-level domain adaptation between synthetic images and real-world ones. The utilization of synthetic data reduces the required amount of real-world samples by up to 50 times.~\cite{mueller2018driving} transfer driving policies trained from simulation to reality via modularity and abstraction, where the driving policy is exposed to segmentation results of input scenes and target way points, instead of raw perceptual input or low-level vehicle dynamics.~\cite{jia2021physics} incorporate the task-specific prior knowledge into the model and pre-train it with synthetic data generated by imperfect physical models, which allows the model to get close enough to the target solution and only a small amount of real-world data is needed for refining.

\subsubsection{Multitask Learning and Meta Learning with Auxiliary Tasks}
Synthetic data generated from physics-based simulators can also be used to construct auxiliary learning tasks for improving the model's performance on the target task with techniques of multitask learning and meta learning.~\cite{schutt2017schnet} constructs the auxiliary task as the prediction of interatomic forces for the main task of molecular energy prediction. The labels of the auxiliary task and the main task are generated from simulation at the same time, while the prediction of the auxiliary task is produced via differentiating the energy prediction model. Both tasks are used to train the model simultaneously.~\cite{thanasutives2021adversarial} adopts the multitask learning scheme by learning shared representations between multiple related PDEs, which are generated by varying coefficients, for better generalizability of the proposed neural network based PDE solver.~\cite{seo2020physics} proposes a spatiotemporal forecasting model with decoupled spatial and temporal modules, where the spatial module is PDE-independent and are trained via model agnostic meta learning (MAML)~\cite{finn2017model} for fast adaptation on new tasks, while the task-dependent temporal module is trained from scratch for each task.

\subsection{Physics-Informed Neural Network Architecture Design}

\subsubsection{Physics-Informed Computation Graph}
A typical way of physics-informed neural network architecture design is to design computation graphs that mimic the behavior of physics knowledge based methods. While the specific method is highly dependent on the physics knowledge, a general idea is to start from some existing physics based solution, then replace difficult-to-estimate variables with outputs of neural networks, or relax some fixed parameters by enabling them to adapt to the data. In following paragraphs, we will introduce several knowledge-specific neural network designs as well as techniques to directly fuse deep learning models with physics based solutions as a hybrid model.

\paragraph{Energy Conservation Laws}
As introduced in Sec~\ref{sec:classical-mechanics-and-energy-conservation-laws}, Lagrangian and Hamiltonian mechanics are powerful in enforcing energy conservation laws, thus a series of recent works develop neural network architectures based on them to incorporate the energy conservation property.~\cite{lutter2018deep} proposes a network topology named Deep Lagrangian Networks (DeLaN) encoding the Lagrange-Euler PDE originating from Lagrangian Mechanics, which can be trained with standard optimizers while maintaining physical plausibility.~\cite{cranmer2020lagrangian} designs Lagrangian Neural Networks (LNNs) to model arbitrary Lagrangian functions via neural networks, and solve the dynamics of the system with a numerical expression derived from the Euler-Lagrangian equation, where gradients from auto differentiation are utilized. Similarly,~\cite{greydanus2019hamiltonian} (HNN) models the Hamiltonian function with a neural network. The derivatives of spatial coordinates and momentum with respect to time are derived from Eq~\ref{eq:hamiltonian-derive}.~\cite{Toth2020Hamiltonian} proposes Neural Hamiltonian Flow (NHF), which is a powerful normalising flow model using Hamiltonian dynamics as the invertible function to model expressive densities. In NHF, the density is decomposed into the "coordinate" part and the "momentum" part in the hidden space, both of which are then propagated with Eq~\ref{eq:hamiltonian-derive}. The propagation is (1) invertible and (2) has the volume ("energy" with respect to the hidden space) preserving property, which satisfies the requirement of normalising flows. Compared to other flow-based approaches, NHF enjoys higher computational efficiency since it avoids the expensive step of calculating the trace of Jacobians.~\cite{zhong2019symplectic} designs the computation graph of the proposed neural network following the Hamiltonian dynamics with control to incorporate the corresponding inductive bias. It further proposes a parameterization that can enforce the Hamiltonian mechanics with coordinates embedded in a high-dimensional space or velocity data instead of momentum.

\paragraph{Symmetry}
Table~\ref{tab:geodl} in Sec~\ref{sec:symmetry} has shown the connection between some widely used neural network architectures and the corresponding symmetry groups. Here we introduce methods incorporating other types of symmetry groups into neural network architectures. Based on the representations of symmetry groups adopted, all the methods we introduce can be categorized to methods using (1) invariant treatment of coordinates (2) irreducible representations or (3) regular representations.

\textbf{Invariant Treatment of Coordinates} Depending on the symmetry group, spatial coordinates should be properly processed instead of being used as raw inputs.~\cite{satorras2021n} develops equivariant message passing to E(3) via letting messages passed among nodes only depend on distances, which follows the property that E(3) preserves distances between nodes. The same technique is also used in ~\cite{schutt2017schnet}.~\cite{horie2021isometric} propose a set of transformation invariant and equivariant GNN models by tweaking the definition of an adjacency matrix named \textit{isometric adjacency matrix}, which can be viewed as a weighted adjacency matrix for each direction and reflects spatial information.

\textbf{Irreducible Representations} All elements of a roto-translation group can be transformed into an irreducible form: a vector that is rotated by a block diagonal matrix. The full set of equivariant mappings for  some symmetry group can be solved with equivariance constraint over convolution kernels. The solutions form a linear combination of equivariant basis matrices, which can be used for equivariant convolutions. ~\cite{weiler2019general} gives a general solution of the kernel space constraint for arbitrary representations of the Euclidean group E(2) and its subgroups, which forms a wide range of equivariant network architectures.~\cite{thomas2018tensor} develops convolution filters locally equivariant to 3D rotations, translations, and permutations, which are built from spherical harmonics.~\cite{fuchs2020se} further enhances~\cite{thomas2018tensor} with the self-attention mechanism.

\textbf{Regular Representations} Regular representation approaches store copies of latent feature embeddings for all elements of a symmetry group. To mitigate this computational burden, some recent works such as ~\cite{finzi2020generalizing,hutchinson2021lietransformer} use \textit{Lie groups} as the tool for rapid prototyping across various symmetry groups. Only the exponential and logarithm maps are required for incorporating equivariance to a new symmetry group.

\paragraph{Numerical Methods}
Sec~\ref{sec:numerical-pde} introduces several numerical methods for solving PDEs. In this paragraph, we introduce some recent works integrating numerical solutions for each method.

\textbf{Finite Difference Method} \cite{long2018pde,long2019pde} propose learnable differential operators by learning convolution kernels to approximate unknown nonlinear responses in PDEs. All kernels are properly constrained by fully exploiting the relation between the orders of differential operators and the orders of sum rules of filters, which originates from wavelet theory. These constraints ensure both the model's ability to identify PDEs and its expressivity.~\cite{jiang2018spherical} proposes an efficient convolution kernel on unstructured grids of spherical signals using parameterized finite difference operators.~\cite{Seo*2020Physics-aware} leverages differences of sparsely available data from physical systems via the spatial difference layer (SDL). SDL is inspired by finite difference operators on graph and triangulated mesh and replaces fixed parameters in these operators with the output of a GNN capturing spatial information of the input data.

\textbf{Finite Volume Method} \cite{bar2019learning} adopts CNNs to generate coefficients for approximating spatial derivatives, followed by the standard finite volume method. CNNs are optimized end-to-end to best satisfy the equations on low resolution grids, and produce accurate numerical results: it can be integrated at resolutions 4-8x coarser than is possible with standard finite volume methods.~\cite{wang2019learning} creates new PDE solvers based on the WENO scheme ~\cite{shu1998essentially} via generating its coefficients from a learned policy network trained with reinforcement learning.

\textbf{Finite Element Method}
~\cite{alet2019graph} mimics the behavior of finite element analysis: it assigns nodes of a GNN to selected spatial locations and uses message passing on the graph to model the relationship between an initial function and a resulting function defined in the same space. Both the locations of nodes and their connectivity can be optimized to focus on the most important parts of the space.~\cite{xue2020amortized} proposes a two-stage optimization framework for PDE-constrained optimization problem. At the first stage, the framework obtains a surrogate model to prediction solutions of finite element method directly from control parameters. At the second stage, the framework performs gradient-based PDE-constrained optimization.~\cite{trask2019gmls} introduces convolution operators on unstructured point clouds based on Generalized Moving Least Squares (GMLS), which is a non-parametric technique in finite element method~\cite{lim2007mls} for estimating linear bounded functionals from scattered data.

\paragraph{Koopman Theory}~\cite{lusch2018deep} first combines deep learning models with the Koopman operator. It utilizes the power of deep learning to identify nonlinear coordinates on which the dynamics are globally linear using the Koopman operator. The resulting method benefits from both the power and generality of deep learning models and the physical interpretability of Koopman embeddings.~\cite{takeishi2017learning} proposes minimization of the residula sum of squares of linear least-squares regression to estimate the encoders and decoders that maps data into the Koopman invariant subspaces where the linear regression fits well.~\cite{li2020learning} extends deep learning based Koopman operators to scenarios with multiple objects. It adopts GNNs to encode object-centric states and uses a block-wise linear state transition matrix (Koopman matrix) to enforce the shared structure among objects.~\cite{azencot2020forecasting} incorporates the consistency by penalizing the consistency mismatch of forward and backward Koopman matrices.

\subsubsection{Fusion of Deep Learning and Physics-Based Modules}
In addition to previously introduced methods that mimic the behaviour of physics based solutions in the computation graph design of neural network layers, deep learning and physics-based methods can also be fused in a higher level: modules constructed with standard neural network blocks and physics rules work interactively but only expose input/output to each other.

\cite{garcia2019combining} adopts a graphical model derived from equations of motion in a physics model to predict the next future state while adding it with the predicted residual part from a GNN module.
\cite{long2018hybridnet} presents HybridNet, a framework combining data-driven deep learning and model-driven computation for reliable spatiotemporal evolution prediction. The deep learning part, Convolutional LSTM (ConvLSTM) works as the backbone to predict the evolution of external input to the system. The model-driven part, Cellular Neural Network (CeNN), transforms numerical computation in PDE solvers and is able to infer unknown physical parameters.
\cite{belbute2020combining} proposes a hybrid approach for fluid flow prediction containing two components: one GNN-based module operating directly on the original fine-grained non-uniform mesh used in CFD, and one CFD solver operating on a much coarser resolution. The output of the CFD solver is upsampled to the fine-grained mesh and then concatenated to the hidden embeddings from GNN layers. The hybrid model generalizes better than pure GNN-based approaches and is still faster than directly running CFD simulation on the original mesh.
\cite{long2018pde,guen2020disentangling} uses constrained convolution kernels to extract approximated spatial derivatives as features, which are fed as the input to the following neural network layers that capture unknown dynamics and give the final prediction. 

\subsection{Physics-Informed Optimization}
Prior physics knowledge can also be integrated into the optimization process in the form of loss functions directly derived from task-specific knowledge or regularizations from physics principles. The integration of physics knowledge in optimization targets reshapes the optimization space and encourages the training process to converge to physical plausible solutions.

\subsubsection{Task-Specific Knowledge Based Loss Terms}
A variety of physics knowledge can be described in the form of PDEs, which provides connections between the spatial and temporal derivatives as well as constraints of values of interest on boundaries.~\cite{raissi2017physicsI,raissi2017physicsII,raissi2018deep,raissi2019physics} adopts multi-layer perceptrons to directly model the mapping from input spatial and temporal coordinates to the value of PDE solutions. Instead of only minimizing the prediction error between the output and solutions from numerical methods, the series of works adds loss terms enforcing the equation structure, which penalizes the violation of PDEs using derivatives from auto-differentiation of the neural network solution.~\cite{esmaeilzadeh2020meshfreeflownet} optimizes the model for super-resolution of turbulent flows by minimizing a weighted combination of two losses: one is the norm of the difference between predictions and ground truth values, the other is the norm of residues of the governing PDEs.~\cite{Shi_Mo_Di_2021} predicts the traffic flow with a neural network taking time and coordinates as input, and constructs physical discrepancy loss terms with an existing second-order traffic model.~\cite{jia2021physics} trains the proposed lake temperature prediction model with generalized loss function to include the physical consistency-based penalty, which encourages the consistency between lake energy and energy fluxes.

\subsubsection{Regularization}
\cite{shi2019neural} adopts a deep neural network to predict the unknown disturbance forces in the controller of drones. To guarantee the system's stability, the authors first derive the overall stability and robustness requirement indicating constraints on the Lipschitz constant, then minimizes its upper bound - the spectral norm of weights in each layer - together with the prediction error in the optimization.~\cite{holl2019learning,yin2021augmenting} both augment purely physics laws/rule-based prediction models with learnable control signals to mitigate the approximation errors of prior knowledge. While they differ in terms of processes for solving PDEs, both have constraints minimizing the norm of control signals, which originates from the least action principle~\cite{feynman2005principle}.

\section{Challenges and Future Directions}
\label{sec:discussion}

\subsection{Challenge 1: Handcrafted Selection of Physics Knowledge for Incorporation}
Existing works require expertise of the domain-specific knowledge of tasks to incorporate the most appropriate physics knowledge. While this serves the purpose of leveraging domain-knowledge to mitigate the deficiencies of pure data-driven methods, it lacks the flexibility of identifying the correct physics knowledge depending on the task. For example, ~\cite{esmaeilzadeh2020meshfreeflownet} directly uses the governing equation of the Rayleigh-Benard instability problem as the prior knowledge for turbulence super-resolution, while ~\cite{wang2020towards} chooses the derived Hybrid RANS(Reynolds-averaged Navier-Stokes)-LES(Large Eddy Simulation) Coupling method for turbulence prediction. Although both super-resolution and prediction tasks are defined on the same turbulence system governed by the same physics laws, choosing the form of physics laws (the original form/the derived approximation form) to incorporate is heuristic.
\paragraph{Research Direction 1: Automatic Identification of Proper Physics Knowledge To Incorporate}
A promising research direction is to reach the middle ground between the domain-specific knowledge and the pure data-driven way. Here we discuss some potential approaches to realizing it.

\textbf{Neural Architecture Search (NAS)}  The development of NAS allows the automatic design of neural network architectures and NAS methods have outperformed manually designed architectures on many tasks including image classification and semantic segmentation~\cite{elsken2019neural}. By restricting the available physics knowledge within a pre-selected search space and enabling the model to discover the optimal knowledge or combination of knowledge using NAS techniques, the resulting architecture can reach the balance between exploiting prior knowledge and adapting to observed data. ~\cite{skomski2021automating} presents a neural block dynamics design space of neural network components that encompasses various state-space models as the search space for NAS, and models given by NAS in such a space achieve highly accuracy with physically consistent results.

\textbf{Automatic Modularization of Network Architectures}  Modularization of network architectures plays a key part in physics-informed network architecture design. For example, DeLaN~\cite{lutter2018deep} and HNN~\cite{greydanus2019hamiltonian} contain modules estimating the Lagrangian/Hamiltonian function of the system and following modules deriving the prediction. ~\cite{bar2019learning,Seo*2020Physics-aware} separate modules approximating spatial and temporal derivatives in the network. Modularity provides better generalization ability~\cite{xu2020can}, and some modules can further be trained in multiple tasks and benefit the overall performance on all tasks~\cite{alet2019neural,seo2020physics}. However, existing works still require a certain selection of physics knowledge to guide the modularization. Some recent works have started exploring discovering functional modules automatically during the training process. For example, ~\cite{chen2020modular} divides models into reusable modules and task-specific modules by estimating the variance of parameters across tasks. ~\cite{goyal2021recurrent} lets multiple groups of recurrent cells compete with each other so that they are only updated at time steps where they are most relevant. This enables the model to learn modular structures and leads to improved generalization.

\subsection{Challenge 2: Lack of Benchmarks and Evaluations of PIML Methods}
Comprehensive benchmarks have shown as great boosters for the development of corresponding research areas. Examples include ImageNet Large Scale Visual Recognition Challenge(ILSVRC)~\cite{ILSVRC15} and Common Objects in Context(COCO)~\cite{lin2014microsoft} from computer vision, Workshop on Statistical Machine Translation (WMT)~\cite{bojar2014findings} and Stanford Question Answering Dataset (SQuAD)~\cite{rajpurkar2016squad} from natural language processing. However, due to the complexity and heterogeneity of problem settings, PIML still lacks comprehensive benchmarks for evaluating various methods of knowledge integration, which creates barriers in the development of PIML. First, most problems in PIML come from physics or engineering applications, where acquiring the data and formalizing the task can be challenging for researchers without domain knowledge and experience. Second, existing works such as ~\cite{lutter2018deep,greydanus2019hamiltonian,Seo*2020Physics-aware,azencot2020forecasting,long2018pde,bar2019learning,belbute2020combining} heavily rely on heterogeneous domain-specific datasets, which greatly increases the difficulty of fairly comparing different PIML methods.
\paragraph{Research Direction 2: Comprehensive Benchmarks for PIML Methods}
Constructing comprehensive benchmarks for PIML is of great need for boosting its development. According to the above discussion, ideal benchmarks  (1) must provide publicly available and organized datasets, and formulate benchmark tasks as typical machine learning tasks, such as classification and regression; (2) must be general enough to accommodate various PIML methods as well as data-driven and pure physics-based methods. Recent development along this direction includes \textbf{WeatherBench}~\cite{rasp2020weatherbench} and \textbf{Open Graph Benchmark (OGB)}~\cite{hu2020open}. WeatherBench provides processed weather data together with clearly defined tasks and evaluation metrics for medium-range weather forecasting. It also presents performance of purely data-driven models and numerical models as baselines. OGB offers data of protein structures and molecule structures, and construct node/link/graph property prediction tasks, where performance of PIML methods and data-driven baselines can be directly compared.

\subsection{Challenge 3: Suboptimal Existing Neural Network Architectures and Optimization Methods for PIML}
Established theories and empirical conclusions of neural network architectures and optimization methods are mostly developed in areas where neural network methods first gain advantages, such as computer vision and natural language processing. However, they may no longer be effective in PIML. The reason is that PIML methods usually involve explicit use of gradients in forwarding processes and objective functions, leading to the existence of high-order derivatives in the backward process, which shapes optimization spaces significantly different from typical deep learning models. For example, \cite{cranmer2020lagrangian} notices that regular parameter initialization methods such as Kaiming~\cite{he2015delving} and Xavier~\cite{glorot2010understanding} are insufficient since the unusual optimization objective is very nonlinear.\cite{jagtap2020adaptive} also empirically demonstrates that the widely used Rectified Linear Unit (ReLU) activation is not effective in the physics-informed PINN architecture proposed by ~\cite{raissi2017physicsI,raissi2017physicsII}.

\paragraph{Research Direction 3: Novel Neural Network Designs for PIML}
The drastic differences in network architecture and objectives between PIML and conventional deep learning tasks signify the importance of novel neural network designs for PIML from both the architecture and the optimization aspects. 

From the architecture perspective, since many PIML methods involve the utilization of gradients from the auto-differentiation of neural networks, designing new architectures/components better preserving gradient information is one promising direction. \cite{sitzmann2020implicit} leverages periodic activation functions for implicit neural representations and demonstrate that they are ideally suited for representing complex natural signals and their spatial/temporal derivatives.

From the optimization perspective, multiple objectives (including both the task objective and the physics-informed constraints) may contradict with each other and lead to suboptimal results under vanilla optimization methods.~\cite{kim2021dpm} notices the discrepancy of updating directions between the boundary constraint loss and the approximation loss in PINN~\cite{raissi2017physicsI,raissi2017physicsII}, and proposes the dynamic pulling method (DPM) to align their updating directions, which significantly improves the extrapolation performance of PINNs.

\section{Summary}
\label{sec:conclusion}
In this paper, we provide a thorough and comprehensive survey of existing works in PIML. We summarize them from three aspects: (1) motivations of PIML, (2) physics knowledge in PIML, (3) methods of knowledge integration in PIML. In the end, we discuss existing challenges of PIML and indicate potential future research directions accordingly. We expect the paper can serve as the guide for PIML users to select proper physics knowledge and appropriate integration methods, as well as the guide for PIML researchers to identify existing gaps and promising research directions.

\bibliographystyle{unsrt}
\bibliography{references}

\begin{thebibliography}{100}

\bibitem{lecun1998gradient}
Yann LeCun, L{\'e}on Bottou, Yoshua Bengio, and Patrick Haffner.
\newblock Gradient-based learning applied to document recognition.
\newblock {\em Proceedings of the IEEE}, 86(11):2278--2324, 1998.

\bibitem{krizhevsky2012imagenet}
Alex Krizhevsky, Ilya Sutskever, and Geoffrey~E Hinton.
\newblock Imagenet classification with deep convolutional neural networks.
\newblock In {\em Advances in neural information processing systems}, pages
  1097--1105, 2012.

\bibitem{he2016deep}
Kaiming He, Xiangyu Zhang, Shaoqing Ren, and Jian Sun.
\newblock Deep residual learning for image recognition.
\newblock In {\em Proceedings of the IEEE conference on computer vision and
  pattern recognition}, pages 770--778, 2016.

\bibitem{redmon2016you}
Joseph Redmon, Santosh Divvala, Ross Girshick, and Ali Farhadi.
\newblock You only look once: Unified, real-time object detection.
\newblock In {\em Proceedings of the IEEE conference on computer vision and
  pattern recognition}, pages 779--788, 2016.

\bibitem{he2017mask}
Kaiming He, Georgia Gkioxari, Piotr Doll{\'a}r, and Ross Girshick.
\newblock Mask r-cnn.
\newblock In {\em Proceedings of the IEEE international conference on computer
  vision}, pages 2961--2969, 2017.

\bibitem{mikolov2013efficient}
Tomas Mikolov, Kai Chen, Greg Corrado, and Jeffrey Dean.
\newblock Efficient estimation of word representations in vector space.
\newblock {\em International Conference on Learning Representations}, 2013.

\bibitem{socher2013recursive}
Richard Socher, Alex Perelygin, Jean Wu, Jason Chuang, Christopher~D Manning,
  Andrew~Y Ng, and Christopher Potts.
\newblock Recursive deep models for semantic compositionality over a sentiment
  treebank.
\newblock In {\em Proceedings of the 2013 conference on empirical methods in
  natural language processing}, pages 1631--1642, 2013.

\bibitem{sutskever2014sequence}
Ilya Sutskever, Oriol Vinyals, and Quoc~V Le.
\newblock Sequence to sequence learning with neural networks.
\newblock {\em Advances in Neural Information Processing Systems}, 2014.

\bibitem{kim2014convolutional}
Yoon Kim.
\newblock Convolutional neural networks for sentence classification.
\newblock In {\em Proceedings of the 2014 Conference on Empirical Methods in
  Natural Language Processing ({EMNLP})}, pages 1746--1751, Doha, Qatar,
  October 2014. Association for Computational Linguistics.

\bibitem{bahdanau2014neural}
Dzmitry Bahdanau, Kyunghyun Cho, and Yoshua Bengio.
\newblock Neural machine translation by jointly learning to align and
  translate.
\newblock {\em arXiv preprint arXiv:1409.0473}, 2014.

\bibitem{kumar2016ask}
Ankit Kumar, Ozan Irsoy, Peter Ondruska, Mohit Iyyer, James Bradbury, Ishaan
  Gulrajani, Victor Zhong, Romain Paulus, and Richard Socher.
\newblock Ask me anything: Dynamic memory networks for natural language
  processing.
\newblock In {\em International conference on machine learning}, pages
  1378--1387. PMLR, 2016.

\bibitem{vaswani2017attention}
Ashish Vaswani, Noam Shazeer, Niki Parmar, Jakob Uszkoreit, Llion Jones,
  Aidan~N Gomez, \L~ukasz Kaiser, and Illia Polosukhin.
\newblock Attention is all you need.
\newblock In I.~Guyon, U.~V. Luxburg, S.~Bengio, H.~Wallach, R.~Fergus,
  S.~Vishwanathan, and R.~Garnett, editors, {\em Advances in Neural Information
  Processing Systems}, volume~30, pages 5998--6008. Curran Associates, Inc.,
  2017.

\bibitem{peters2018deep}
Matthew~E Peters, Mark Neumann, Mohit Iyyer, Matt Gardner, Christopher Clark,
  Kenton Lee, and Luke Zettlemoyer.
\newblock Deep contextualized word representations.
\newblock {\em arXiv preprint arXiv:1802.05365}, 2018.

\bibitem{devlin2019bert}
Jacob Devlin, Ming-Wei Chang, Kenton Lee, and Kristina Toutanova.
\newblock {BERT}: Pre-training of deep bidirectional transformers for language
  understanding.
\newblock In {\em Proceedings of the 2019 Conference of the North {A}merican
  Chapter of the Association for Computational Linguistics: Human Language
  Technologies, Volume 1 (Long and Short Papers)}, pages 4171--4186,
  Minneapolis, Minnesota, June 2019. Association for Computational Linguistics.

\bibitem{hsieh2009machine}
William~W Hsieh.
\newblock {\em Machine learning methods in the environmental sciences: Neural
  networks and kernels}.
\newblock Cambridge university press, 2009.

\bibitem{ivezic2014statistics}
{\v{Z}}eljko Ivezi{\'c}, Andrew~J Connolly, Jacob~T VanderPlas, and Alexander
  Gray.
\newblock {\em Statistics, data mining, and machine learning in astronomy: a
  practical Python guide for the analysis of survey data}, volume~1.
\newblock Princeton University Press, 2014.

\bibitem{karpatne2017theory}
Anuj Karpatne, Gowtham Atluri, James~H Faghmous, Michael Steinbach, Arindam
  Banerjee, Auroop Ganguly, Shashi Shekhar, Nagiza Samatova, and Vipin Kumar.
\newblock Theory-guided data science: A new paradigm for scientific discovery
  from data.
\newblock {\em IEEE Transactions on knowledge and data engineering},
  29(10):2318--2331, 2017.

\bibitem{karpatne2018machine}
Anuj Karpatne, Imme Ebert-Uphoff, Sai Ravela, Hassan~Ali Babaie, and Vipin
  Kumar.
\newblock Machine learning for the geosciences: Challenges and opportunities.
\newblock {\em IEEE Transactions on Knowledge and Data Engineering},
  31(8):1544--1554, 2018.

\bibitem{kutz2017deep}
J~Nathan Kutz.
\newblock Deep learning in fluid dynamics.
\newblock {\em Journal of Fluid Mechanics}, 814:1--4, 2017.

\bibitem{reichstein2019deep}
Markus Reichstein, Gustau Camps-Valls, Bjorn Stevens, Martin Jung, Joachim
  Denzler, Nuno Carvalhais, et~al.
\newblock Deep learning and process understanding for data-driven earth system
  science.
\newblock {\em Nature}, 566(7743):195--204, 2019.

\bibitem{wang2018successful}
Zhen Wang, Haibin Di, Muhammad~Amir Shafiq, Yazeed Alaudah, and Ghassan
  AlRegib.
\newblock Successful leveraging of image processing and machine learning in
  seismic structural interpretation: A review.
\newblock {\em The Leading Edge}, 37(6):451--461, 2018.

\bibitem{von2019informed}
Laura von Rueden, Sebastian Mayer, Katharina Beckh, Bogdan Georgiev, Sven
  Giesselbach, Raoul Heese, Birgit Kirsch, Julius Pfrommer, Annika Pick,
  Rajkumar Ramamurthy, et~al.
\newblock Informed machine learning--a taxonomy and survey of integrating
  knowledge into learning systems.
\newblock {\em arXiv preprint arXiv:1903.12394}, 2019.

\bibitem{esmaeilzadeh2020meshfreeflownet}
Soheil Esmaeilzadeh, Kamyar Azizzadenesheli, Karthik Kashinath, Mustafa
  Mustafa, Hamdi~A Tchelepi, Philip Marcus, Mr~Prabhat, Anima Anandkumar,
  et~al.
\newblock Meshfreeflownet: a physics-constrained deep continuous space-time
  super-resolution framework.
\newblock In {\em SC20: International Conference for High Performance
  Computing, Networking, Storage and Analysis}, pages 1--15. IEEE, 2020.

\bibitem{wang2020towards}
Rui Wang, Karthik Kashinath, Mustafa Mustafa, Adrian Albert, and Rose Yu.
\newblock Towards physics-informed deep learning for turbulent flow prediction.
\newblock In {\em Proceedings of the 26th ACM SIGKDD International Conference
  on Knowledge Discovery \& Data Mining}, pages 1457--1466, 2020.

\bibitem{kashinath2020enforcing}
Karthik Kashinath, Philip Marcus, et~al.
\newblock Enforcing physical constraints in cnns through differentiable pde
  layer.
\newblock In {\em ICLR 2020 Workshop on Integration of Deep Neural Models and
  Differential Equations}, 2020.

\bibitem{rasp2020weatherbench}
Stephan Rasp, Peter~D Dueben, Sebastian Scher, Jonathan~A Weyn, Soukayna
  Mouatadid, and Nils Thuerey.
\newblock Weatherbench: a benchmark data set for data-driven weather
  forecasting.
\newblock {\em Journal of Advances in Modeling Earth Systems},
  12(11):e2020MS002203, 2020.

\bibitem{weyn2020improving}
Jonathan~A Weyn, Dale~R Durran, and Rich Caruana.
\newblock Improving data-driven global weather prediction using deep
  convolutional neural networks on a cubed sphere.
\newblock {\em Journal of Advances in Modeling Earth Systems},
  12(9):e2020MS002109, 2020.

\bibitem{gronquist2021deep}
Peter Gr{\"o}nquist, Chengyuan Yao, Tal Ben-Nun, Nikoli Dryden, Peter Dueben,
  Shigang Li, and Torsten Hoefler.
\newblock Deep learning for post-processing ensemble weather forecasts.
\newblock {\em Philosophical Transactions of the Royal Society A},
  379(2194):20200092, 2021.

\bibitem{kashinath2021physics}
K~Kashinath, M~Mustafa, A~Albert, JL~Wu, C~Jiang, S~Esmaeilzadeh,
  K~Azizzadenesheli, R~Wang, A~Chattopadhyay, A~Singh, et~al.
\newblock Physics-informed machine learning: case studies for weather and
  climate modelling.
\newblock {\em Philosophical Transactions of the Royal Society A},
  379(2194):20200093, 2021.

\bibitem{sanchez2020learning}
Alvaro Sanchez-Gonzalez, Jonathan Godwin, Tobias Pfaff, Rex Ying, Jure
  Leskovec, and Peter Battaglia.
\newblock Learning to simulate complex physics with graph networks.
\newblock In {\em International Conference on Machine Learning}, pages
  8459--8468. PMLR, 2020.

\bibitem{li2018learning}
Yunzhu Li, Jiajun Wu, Russ Tedrake, Joshua~B. Tenenbaum, and Antonio Torralba.
\newblock Learning particle dynamics for manipulating rigid bodies, deformable
  objects, and fluids.
\newblock In {\em International Conference on Learning Representations}, 2019.

\bibitem{ummenhofer2019lagrangian}
Benjamin Ummenhofer, Lukas Prantl, Nils Thuerey, and Vladlen Koltun.
\newblock Lagrangian fluid simulation with continuous convolutions.
\newblock In {\em International Conference on Learning Representations}, 2019.

\bibitem{pfaff2021learning}
Tobias Pfaff, Meire Fortunato, Alvaro Sanchez-Gonzalez, and Peter Battaglia.
\newblock Learning mesh-based simulation with graph networks.
\newblock In {\em International Conference on Learning Representations}, 2021.

\bibitem{belbute2020combining}
Filipe de~Avila Belbute-Peres, Thomas Economon, and Zico Kolter.
\newblock Combining differentiable pde solvers and graph neural networks for
  fluid flow prediction.
\newblock In {\em International Conference on Machine Learning}, pages
  2402--2411. PMLR, 2020.

\bibitem{crutchfield1987equations}
James~P Crutchfield and BS~McNamara.
\newblock Equations of motion from a data series.
\newblock {\em Complex systems}, 1(417-452):121, 1987.

\bibitem{kevrekidis2003equation}
Ioannis~G Kevrekidis, C~William Gear, James~M Hyman, Panagiotis~G Kevrekidid,
  Olof Runborg, Constantinos Theodoropoulos, et~al.
\newblock Equation-free, coarse-grained multiscale computation: Enabling
  mocroscopic simulators to perform system-level analysis.
\newblock {\em Communications in Mathematical Sciences}, 1(4):715--762, 2003.

\bibitem{raissi2017physicsI}
Maziar Raissi, Paris Perdikaris, and George~Em Karniadakis.
\newblock Physics informed deep learning (part i): Data-driven solutions of
  nonlinear partial differential equations.
\newblock {\em arXiv preprint arXiv:1711.10561}, 2017.

\bibitem{raissi2017physicsII}
Maziar Raissi, Paris Perdikaris, and George~Em Karniadakis.
\newblock Physics informed deep learning (part ii): Data-driven discovery of
  nonlinear partial differential equations.
\newblock {\em arXiv preprint arXiv:1711.10566}, 2017.

\bibitem{raissi2018deep}
Maziar Raissi.
\newblock Deep hidden physics models: Deep learning of nonlinear partial
  differential equations.
\newblock {\em arXiv preprint arXiv:1801.06637}, 2018.

\bibitem{magill2018neural}
Martin Magill, Faisal Qureshi, and Hendrick~W de~Haan.
\newblock Neural networks trained to solve differential equations learn general
  representations.
\newblock {\em Advances in Neural Information Processing Systems}, 2018.

\bibitem{li2020neural}
Zongyi Li, Nikola Kovachki, Kamyar Azizzadenesheli, Burigede Liu, Kaushik
  Bhattacharya, Andrew Stuart, and Anima Anandkumar.
\newblock Neural operator: Graph kernel network for partial differential
  equations.
\newblock {\em arXiv preprint arXiv:2003.03485}, 2020.

\bibitem{li2021fourier}
Zongyi Li, Nikola~Borislavov Kovachki, Kamyar Azizzadenesheli, Kaushik
  Bhattacharya, Andrew Stuart, Anima Anandkumar, et~al.
\newblock Fourier neural operator for parametric partial differential
  equations.
\newblock In {\em International Conference on Learning Representations}, 2021.

\bibitem{bar2019learning}
Yohai Bar-Sinai, Stephan Hoyer, Jason Hickey, and Michael~P Brenner.
\newblock Learning data-driven discretizations for partial differential
  equations.
\newblock {\em Proceedings of the National Academy of Sciences},
  116(31):15344--15349, 2019.

\bibitem{um2020sol}
Kiwon Um, Robert Brand, Yun Fei, Philipp Holl, and Nils Thuerey.
\newblock {Solver-in-the-Loop: Learning from Differentiable Physics to Interact
  with Iterative PDE-Solvers}.
\newblock {\em Advances in Neural Information Processing Systems}, 2020.

\bibitem{sharifi2019downscaling}
Ehsan Sharifi, B~Saghafian, and R~Steinacker.
\newblock Downscaling satellite precipitation estimates with multiple linear
  regression, artificial neural networks, and spline interpolation techniques.
\newblock {\em Journal of Geophysical Research: Atmospheres}, 124(2):789--805,
  2019.

\bibitem{vandal2017deepsd}
Thomas Vandal, Evan Kodra, Sangram Ganguly, Andrew Michaelis, Ramakrishna
  Nemani, and Auroop~R Ganguly.
\newblock Deepsd: Generating high resolution climate change projections through
  single image super-resolution.
\newblock In {\em Proceedings of the 23rd acm sigkdd international conference
  on knowledge discovery and data mining}, pages 1663--1672, 2017.

\bibitem{lee2020coarse}
Seungjoon Lee, Mahdi Kooshkbaghi, Konstantinos Spiliotis, Constantinos~I
  Siettos, and Ioannis~G Kevrekidis.
\newblock Coarse-scale pdes from fine-scale observations via machine learning.
\newblock {\em Chaos: An Interdisciplinary Journal of Nonlinear Science},
  30(1):013141, 2020.

\bibitem{chan2017parametrization}
Shing Chan and Ahmed~H Elsheikh.
\newblock Parametrization and generation of geological models with generative
  adversarial networks.
\newblock {\em arXiv preprint arXiv:1708.01810}, 2017.

\bibitem{arjovsky2017wasserstein}
Martin Arjovsky, Soumith Chintala, and L{\'e}on Bottou.
\newblock Wasserstein generative adversarial networks.
\newblock In {\em International conference on machine learning}, pages
  214--223. PMLR, 2017.

\bibitem{goldstein2014data}
EB~Goldstein, G~Coco, AB~Murray, and MO~Green.
\newblock Data-driven components in a model of inner-shelf sorted bedforms: a
  new hybrid model.
\newblock {\em Earth Surface Dynamics}, 2(1):67--82, 2014.

\bibitem{brenowitz2018prognostic}
Noah~D Brenowitz and Christopher~S Bretherton.
\newblock Prognostic validation of a neural network unified physics
  parameterization.
\newblock {\em Geophysical Research Letters}, 45(12):6289--6298, 2018.

\bibitem{gentine2018could}
Pierre Gentine, Mike Pritchard, Stephan Rasp, Gael Reinaudi, and Galen Yacalis.
\newblock Could machine learning break the convection parameterization
  deadlock?
\newblock {\em Geophysical Research Letters}, 45(11):5742--5751, 2018.

\bibitem{gilmer2017neural}
Justin Gilmer, Samuel~S Schoenholz, Patrick~F Riley, Oriol Vinyals, and
  George~E Dahl.
\newblock Neural message passing for quantum chemistry.
\newblock In {\em Proceedings of the 34th International Conference on Machine
  Learning-Volume 70}, pages 1263--1272. JMLR. org, 2017.

\bibitem{wang2019molecule}
Xiaofeng Wang, Zhen Li, Mingjian Jiang, Shuang Wang, Shugang Zhang, and
  Zhiqiang Wei.
\newblock Molecule property prediction based on spatial graph embedding.
\newblock {\em Journal of chemical information and modeling}, 59(9):3817--3828,
  2019.

\bibitem{koopman1931hamiltonian}
Bernard~O Koopman.
\newblock Hamiltonian systems and transformation in hilbert space.
\newblock {\em Proceedings of the national academy of sciences of the united
  states of america}, 17(5):315, 1931.

\bibitem{schmid2010dynamic}
Peter~J Schmid.
\newblock Dynamic mode decomposition of numerical and experimental data.
\newblock {\em Journal of fluid mechanics}, 656:5--28, 2010.

\bibitem{williams2015data}
Matthew~O Williams, Ioannis~G Kevrekidis, and Clarence~W Rowley.
\newblock A data--driven approximation of the koopman operator: Extending
  dynamic mode decomposition.
\newblock {\em Journal of Nonlinear Science}, 25(6):1307--1346, 2015.

\bibitem{kevrekidis2016kernel}
I~Kevrekidis, Clarence~W Rowley, and M~Williams.
\newblock A kernel-based method for data-driven koopman spectral analysis.
\newblock {\em Journal of Computational Dynamics}, 2(2):247--265, 2016.

\bibitem{li2020learning}
Yunzhu Li, Hao He, Jiajun Wu, Dina Katabi, and Antonio Torralba.
\newblock Learning compositional koopman operators for model-based control.
\newblock In {\em International Conference on Learning Representations}, 2020.

\bibitem{azencot2020forecasting}
Omri Azencot, N~Benjamin Erichson, Vanessa Lin, and Michael Mahoney.
\newblock Forecasting sequential data using consistent koopman autoencoders.
\newblock In {\em International Conference on Machine Learning}, pages
  475--485. PMLR, 2020.

\bibitem{lusch2018deep}
Bethany Lusch, J~Nathan Kutz, and Steven~L Brunton.
\newblock Deep learning for universal linear embeddings of nonlinear dynamics.
\newblock {\em Nature communications}, 9(1):1--10, 2018.

\bibitem{xiao2019reduced}
D~Xiao, CE~Heaney, L~Mottet, F~Fang, W~Lin, IM~Navon, Y~Guo, OK~Matar,
  AG~Robins, and CC~Pain.
\newblock A reduced order model for turbulent flows in the urban environment
  using machine learning.
\newblock {\em Building and Environment}, 148:323--337, 2019.

\bibitem{mohan2018deep}
Arvind~T Mohan and Datta~V Gaitonde.
\newblock A deep learning based approach to reduced order modeling for
  turbulent flow control using lstm neural networks.
\newblock {\em arXiv preprint arXiv:1804.09269}, 2018.

\bibitem{rubin1974estimating}
Donald~B Rubin.
\newblock Estimating causal effects of treatments in randomized and
  nonrandomized studies.
\newblock {\em Journal of educational Psychology}, 66(5):688, 1974.

\bibitem{pearl2009causality}
Judea Pearl.
\newblock {\em Causality}.
\newblock Cambridge university press, 2009.

\bibitem{imbens2015causal}
Guido~W Imbens and Donald~B Rubin.
\newblock {\em Causal inference in statistics, social, and biomedical
  sciences}.
\newblock Cambridge University Press, 2015.

\bibitem{koller2009probabilistic}
Daphne Koller and Nir Friedman.
\newblock {\em Probabilistic graphical models: principles and techniques}.
\newblock MIT press, 2009.

\bibitem{granger1969investigating}
Clive~WJ Granger.
\newblock Investigating causal relations by econometric models and
  cross-spectral methods.
\newblock {\em Econometrica: journal of the Econometric Society}, pages
  424--438, 1969.

\bibitem{runge2018causal}
Jakob Runge.
\newblock Causal network reconstruction from time series: From theoretical
  assumptions to practical estimation.
\newblock {\em Chaos: An Interdisciplinary Journal of Nonlinear Science},
  28(7):075310, 2018.

\bibitem{runge2019detecting}
Jakob Runge, Peer Nowack, Marlene Kretschmer, Seth Flaxman, and Dino
  Sejdinovic.
\newblock Detecting and quantifying causal associations in large nonlinear time
  series datasets.
\newblock {\em Science Advances}, 5(11):eaau4996, 2019.

\bibitem{runge2019inferring}
Jakob Runge, Sebastian Bathiany, Erik Bollt, Gustau Camps-Valls, Dim Coumou,
  Ethan Deyle, Clark Glymour, Marlene Kretschmer, Miguel~D Mahecha, Jordi
  Mu{\~n}oz-Mar{\'\i}, et~al.
\newblock Inferring causation from time series in earth system sciences.
\newblock {\em Nature communications}, 10(1):1--13, 2019.

\bibitem{nauta2019causal}
Meike Nauta, Doina Bucur, and Christin Seifert.
\newblock Causal discovery with attention-based convolutional neural networks.
\newblock {\em Machine Learning and Knowledge Extraction}, 1(1):312--340, 2019.

\bibitem{pamfil2020dynotears}
Roxana Pamfil, Nisara Sriwattanaworachai, Shaan Desai, Philip Pilgerstorfer,
  Konstantinos Georgatzis, Paul Beaumont, and Bryon Aragam.
\newblock Dynotears: Structure learning from time-series data.
\newblock In {\em International Conference on Artificial Intelligence and
  Statistics}, pages 1595--1605. PMLR, 2020.

\bibitem{zheng2018dags}
Xun Zheng, Bryon Aragam, Pradeep~K Ravikumar, and Eric~P Xing.
\newblock Dags with no tears: Continuous optimization for structure learning.
\newblock In S.~Bengio, H.~Wallach, H.~Larochelle, K.~Grauman, N.~Cesa-Bianchi,
  and R.~Garnett, editors, {\em Advances in Neural Information Processing
  Systems}, volume~31, pages 9472--9483. Curran Associates, Inc., 2018.

\bibitem{hernan2010causal}
Miguel~A Hern{\'a}n and James~M Robins.
\newblock Causal inference, 2010.

\bibitem{robins2000marginal}
James~M Robins, Miguel~Angel Hernan, and Babette Brumback.
\newblock Marginal structural models and causal inference in epidemiology,
  2000.

\bibitem{fitzmaurice2008estimation}
Garrett Fitzmaurice, Marie Davidian, Geert Verbeke, and Geert Molenberghs.
\newblock Estimation of the causal effects of time-varying exposures.
\newblock In {\em Longitudinal Data Analysis}, pages 567--614. Chapman and
  Hall/CRC, 2008.

\bibitem{lim2018forecasting}
Bryan Lim, Ahmed Alaa, and Mihaela van~der Schaar.
\newblock Forecasting treatment responses over time using recurrent marginal
  structural networks.
\newblock {\em NeurIPS}, 18:7483--7493, 2018.

\bibitem{schulam2017reliable}
Peter Schulam and Suchi Saria.
\newblock Reliable decision support using counterfactual models.
\newblock {\em Advances in Neural Information Processing Systems},
  30:1697--1708, 2017.

\bibitem{soleimani2017treatment}
Hossein Soleimani, Adarsh Subbaswamy, and Suchi Saria.
\newblock Treatment-response models for counterfactual reasoning with
  continuous-time, continuous-valued interventions.
\newblock {\em arXiv preprint arXiv:1704.02038}, 2017.

\bibitem{pearl2012measurement}
Judea Pearl.
\newblock On measurement bias in causal inference.
\newblock {\em arXiv preprint arXiv:1203.3504}, 2012.

\bibitem{kuroki2014measurement}
Manabu Kuroki and Judea Pearl.
\newblock Measurement bias and effect restoration in causal inference.
\newblock {\em Biometrika}, 101(2):423--437, 2014.

\bibitem{bica2020tsd}
Ioana Bica, Ahmed~M Alaa, and Mihaela van~der Schaar.
\newblock Time series deconfounder: Estimating treatment effects over time in
  the presence of hidden confounders.
\newblock {\em International Conference on Machine Learning}, 2020.

\bibitem{hatt2021sequential}
Tobias Hatt and Stefan Feuerriegel.
\newblock Sequential deconfounding for causal inference with unobserved
  confounders.
\newblock {\em arXiv preprint arXiv:2104.09323}, 2021.

\bibitem{kuzmanovic2021deconfounding}
Milan Kuzmanovic, Tobias Hatt, and Stefan Feuerriegel.
\newblock Deconfounding temporal autoencoder: estimating treatment effects over
  time using noisy proxies.
\newblock In {\em Machine Learning for Health}, pages 143--155. PMLR, 2021.

\bibitem{liu2020estimating}
Ruoqi Liu, Changchang Yin, and Ping Zhang.
\newblock Estimating individual treatment effects with time-varying
  confounders.
\newblock In {\em 2020 IEEE International Conference on Data Mining (ICDM)},
  pages 382--391. IEEE, 2020.

\bibitem{ma2021deconfounding}
Jing Ma, Ruocheng Guo, Chen Chen, Aidong Zhang, and Jundong Li.
\newblock Deconfounding with networked observational data in a dynamic
  environment.
\newblock In {\em Proceedings of the 14th ACM International Conference on Web
  Search and Data Mining}, pages 166--174, 2021.

\bibitem{battaglia2016interaction}
Peter Battaglia, Razvan Pascanu, Matthew Lai, Danilo~Jimenez Rezende, et~al.
\newblock Interaction networks for learning about objects, relations and
  physics.
\newblock In {\em Advances in neural information processing systems}, pages
  4502--4510, 2016.

\bibitem{chang2016compositional}
Michael~B Chang, Tomer Ullman, Antonio Torralba, and Joshua~B Tenenbaum.
\newblock A compositional object-based approach to learning physical dynamics.
\newblock {\em arXiv preprint arXiv:1612.00341}, 2016.

\bibitem{wu2015galileo}
Jiajun Wu, Ilker Yildirim, Joseph~J Lim, Bill Freeman, and Josh Tenenbaum.
\newblock Galileo: Perceiving physical object properties by integrating a
  physics engine with deep learning.
\newblock {\em Advances in neural information processing systems}, 28, 2015.

\bibitem{levine2014learning}
Sergey Levine and Pieter Abbeel.
\newblock Learning neural network policies with guided policy search under
  unknown dynamics.
\newblock {\em Advances in neural information processing systems}, 27, 2014.

\bibitem{li2019multi}
Minne Li, Lisheng Wu, Jun Wang, and Haitham Bou~Ammar.
\newblock Multi-view reinforcement learning.
\newblock {\em Advances in neural information processing systems}, 32, 2019.

\bibitem{wahlstrom2015pixels}
Niklas Wahlstr{\"o}m, Thomas~B Sch{\"o}n, and Marc~Peter Deisenroth.
\newblock From pixels to torques: Policy learning with deep dynamical models.
\newblock {\em arXiv preprint arXiv:1502.02251}, 2015.

\bibitem{watter2015embed}
Manuel Watter, Jost Springenberg, Joschka Boedecker, and Martin Riedmiller.
\newblock Embed to control: A locally linear latent dynamics model for control
  from raw images.
\newblock {\em Advances in neural information processing systems}, 28, 2015.

\bibitem{he2016opponent}
He~He, Jordan Boyd-Graber, Kevin Kwok, and Hal Daum{\'e}~III.
\newblock Opponent modeling in deep reinforcement learning.
\newblock In {\em International conference on machine learning}, pages
  1804--1813. PMLR, 2016.

\bibitem{tian2019regularized}
Zheng Tian, Ying Wen, Zhichen Gong, Faiz Punakkath, Shihao Zou, and Jun Wang.
\newblock A regularized opponent model with maximum entropy objective.
\newblock {\em arXiv preprint arXiv:1905.08087}, 2019.

\bibitem{Yi*2020CLEVRER:}
Kexin Yi*, Chuang Gan*, Yunzhu Li, Pushmeet Kohli, Jiajun Wu, Antonio Torralba,
  and Joshua~B. Tenenbaum.
\newblock Clevrer: Collision events for video representation and reasoning.
\newblock In {\em International Conference on Learning Representations}, 2020.

\bibitem{phys101}
Jiajun Wu, Joseph~J Lim, Hongyi Zhang, Joshua~B Tenenbaum, and William~T
  Freeman.
\newblock Physics 101: Learning physical object properties from unlabeled
  videos.
\newblock In {\em British Machine Vision Conference}, 2016.

\bibitem{li2020causal}
Yunzhu Li, Antonio Torralba, Anima Anandkumar, Dieter Fox, and Animesh Garg.
\newblock Causal discovery in physical systems from videos.
\newblock {\em Advances in Neural Information Processing Systems},
  33:9180--9192, 2020.

\bibitem{Baradel2020CoPhy:}
Fabien Baradel, Natalia Neverova, Julien Mille, Greg Mori, and Christian Wolf.
\newblock Cophy: Counterfactual learning of physical dynamics.
\newblock In {\em International Conference on Learning Representations}, 2020.

\bibitem{villegas17mcnet}
Ruben Villegas, Jimei Yang, Seunghoon Hong, Xunyu Lin, and Honglak Lee.
\newblock Decomposing motion and content for natural video sequence prediction.
\newblock {\em ICLR}, 2017.

\bibitem{NIPS2017_2d2ca7ee}
Emily~L Denton and vighnesh Birodkar.
\newblock Unsupervised learning of disentangled representations from video.
\newblock In I.~Guyon, U.~V. Luxburg, S.~Bengio, H.~Wallach, R.~Fergus,
  S.~Vishwanathan, and R.~Garnett, editors, {\em Advances in Neural Information
  Processing Systems}, volume~30. Curran Associates, Inc., 2017.

\bibitem{villegas2017learning}
Ruben Villegas, Jimei Yang, Yuliang Zou, Sungryull Sohn, Xunyu Lin, and Honglak
  Lee.
\newblock Learning to generate long-term future via hierarchical prediction.
\newblock In {\em international conference on machine learning}, pages
  3560--3569. PMLR, 2017.

\bibitem{guen2020disentangling}
Vincent~Le Guen and Nicolas Thome.
\newblock Disentangling physical dynamics from unknown factors for unsupervised
  video prediction.
\newblock In {\em Proceedings of the IEEE/CVF Conference on Computer Vision and
  Pattern Recognition}, pages 11474--11484, 2020.

\bibitem{ye2018interpretable}
Tian Ye, Xiaolong Wang, James Davidson, and Abhinav Gupta.
\newblock Interpretable intuitive physics model.
\newblock In {\em Proceedings of the European Conference on Computer Vision
  (ECCV)}, pages 87--102, 2018.

\bibitem{duan2021pip}
Jiafei Duan, Samson Yu, Soujanya Poria, Bihan Wen, and Cheston Tan.
\newblock Pip: Physical interaction prediction via mental imagery with span
  selection.
\newblock {\em arXiv preprint arXiv:2109.04683}, 2021.

\bibitem{janny2022filteredcophy}
Steeven JANNY, Fabien Baradel, Natalia Neverova, Madiha Nadri, Greg Mori, and
  Christian Wolf.
\newblock Filtered-cophy: Unsupervised learning of counterfactual physics in
  pixel space.
\newblock In {\em International Conference on Learning Representations}, 2022.

\bibitem{hornik1989multilayer}
Kurt Hornik, Maxwell Stinchcombe, and Halbert White.
\newblock Multilayer feedforward networks are universal approximators.
\newblock {\em Neural networks}, 2(5):359--366, 1989.

\bibitem{greydanus2019hamiltonian}
Samuel Greydanus, Misko Dzamba, and Jason Yosinski.
\newblock Hamiltonian neural networks.
\newblock In {\em Advances in Neural Information Processing Systems}, pages
  15353--15363, 2019.

\bibitem{kipf2018neural}
Thomas Kipf, Ethan Fetaya, Kuan-Chieh Wang, Max Welling, and Richard Zemel.
\newblock Neural relational inference for interacting systems.
\newblock {\em International Conference on Machine Learning}, 2018.

\bibitem{li2020generative}
Guangyu Li, Bo~Jiang, Hao Zhu, Zhengping Che, and Yan Liu.
\newblock Generative attention networks for multi-agent behavioral modeling.
\newblock In {\em Proceedings of the AAAI Conference on Artificial
  Intelligence}, volume~34, pages 7195--7202, 2020.

\bibitem{yan2018spatial}
Sijie Yan, Yuanjun Xiong, and Dahua Lin.
\newblock Spatial temporal graph convolutional networks for skeleton-based
  action recognition.
\newblock In {\em Thirty-second AAAI conference on artificial intelligence},
  2018.

\bibitem{battaglia2018relational}
Peter~W Battaglia, Jessica~B Hamrick, Victor Bapst, Alvaro Sanchez-Gonzalez,
  Vinicius Zambaldi, Mateusz Malinowski, Andrea Tacchetti, David Raposo, Adam
  Santoro, Ryan Faulkner, et~al.
\newblock Relational inductive biases, deep learning, and graph networks.
\newblock {\em arXiv preprint arXiv:1806.01261}, 2018.

\bibitem{sanchez2018graph}
Alvaro Sanchez-Gonzalez, Nicolas Heess, Jost~Tobias Springenberg, Josh Merel,
  Martin Riedmiller, Raia Hadsell, and Peter Battaglia.
\newblock Graph networks as learnable physics engines for inference and
  control.
\newblock {\em International Conference on Machine Learning}, 2018.

\bibitem{xu2020can}
Keylu Xu, Jingling Li, Mozhi Zhang, Simon~S Du, Ken-ichi Kawarabayashi, and
  Stefanie Jegelka.
\newblock What can neural networks reason about?
\newblock In {\em ICLR}, 2020.

\bibitem{schutt2017schnet}
KT~Sch{\"u}tt, P-J Kindermans, Huziel~E Sauceda, S~Chmiela, Alexandre
  Tkatchenko, and Klaus-Robert M{\"u}ller.
\newblock Schnet: A continuous-filter convolutional neural network for modeling
  quantum interactions.
\newblock In {\em 31st Conference on Neural Information Processing Systems
  (NIPS 2017)}, pages 992--1002, 2018.

\bibitem{lutter2018deep}
Michael Lutter, Christian Ritter, and Jan Peters.
\newblock Deep lagrangian networks: Using physics as model prior for deep
  learning.
\newblock In {\em International Conference on Learning Representations}, 2019.

\bibitem{cranmer2020lagrangian}
Miles Cranmer, Sam Greydanus, Stephan Hoyer, Peter Battaglia, David Spergel,
  and Shirley Ho.
\newblock Lagrangian neural networks.
\newblock In {\em ICLR 2020 Workshop on Integration of Deep Neural Models and
  Differential Equations}, 2020.

\bibitem{li2018dcrnn_traffic}
Yaguang Li, Rose Yu, Cyrus Shahabi, and Yan Liu.
\newblock Diffusion convolutional recurrent neural network: Data-driven traffic
  forecasting.
\newblock In {\em International Conference on Learning Representations (ICLR
  '18)}, 2018.

\bibitem{zhou2021informer}
Haoyi Zhou, Shanghang Zhang, Jieqi Peng, Shuai Zhang, Jianxin Li, Hui Xiong,
  and Wancai Zhang.
\newblock Informer: Beyond efficient transformer for long sequence time-series
  forecasting.
\newblock In {\em Proceedings of AAAI}, 2021.

\bibitem{wu2020connecting}
Zonghan Wu, Shirui Pan, Guodong Long, Jing Jiang, Xiaojun Chang, and Chengqi
  Zhang.
\newblock Connecting the dots: Multivariate time series forecasting with graph
  neural networks.
\newblock In {\em Proceedings of the 26th ACM SIGKDD International Conference
  on Knowledge Discovery \& Data Mining}, pages 753--763, 2020.

\bibitem{He2020AdvectiveNet:}
Xingzhe He, Helen~Lu Cao, and Bo~Zhu.
\newblock Advectivenet: An eulerian-lagrangian fluidic reservoir for point
  cloud processing.
\newblock In {\em International Conference on Learning Representations}, 2020.

\bibitem{Seo*2020Physics-aware}
Sungyong Seo, Chuizheng Meng, and Yan Liu.
\newblock Physics-aware difference graph networks for sparsely-observed
  dynamics.
\newblock In {\em International Conference on Learning Representations}, 2020.

\bibitem{iakovlev2021learning}
Valerii Iakovlev, Markus Heinonen, and Harri L{\"a}hdesm{\"a}ki.
\newblock Learning continuous-time {\{}pde{\}}s from sparse data with graph
  neural networks.
\newblock In {\em International Conference on Learning Representations}, 2021.

\bibitem{Defferrard2020DeepSphere}
Michaël Defferrard, Martino Milani, Frédérick Gusset, and Nathanaël
  Perraudin.
\newblock Deepsphere: a graph-based spherical cnn.
\newblock In {\em International Conference on Learning Representations}, 2020.

\bibitem{armeni2017joint}
Iro Armeni, Sasha Sax, Amir~R Zamir, and Silvio Savarese.
\newblock Joint 2d-3d-semantic data for indoor scene understanding.
\newblock {\em arXiv preprint arXiv:1702.01105}, 2017.

\bibitem{bogo2014faust}
Federica Bogo, Javier Romero, Matthew Loper, and Michael~J Black.
\newblock Faust: Dataset and evaluation for 3d mesh registration.
\newblock In {\em Proceedings of the IEEE Conference on Computer Vision and
  Pattern Recognition}, pages 3794--3801, 2014.

\bibitem{masci2015geodesic}
Jonathan Masci, Davide Boscaini, Michael Bronstein, and Pierre Vandergheynst.
\newblock Geodesic convolutional neural networks on riemannian manifolds.
\newblock In {\em Proceedings of the IEEE international conference on computer
  vision workshops}, pages 37--45, 2015.

\bibitem{monti2017geometric}
Federico Monti, Davide Boscaini, Jonathan Masci, Emanuele Rodola, Jan Svoboda,
  and Michael~M Bronstein.
\newblock Geometric deep learning on graphs and manifolds using mixture model
  cnns.
\newblock In {\em Proc. CVPR}, 2017.

\bibitem{boscaini2016anisotropic}
Davide Boscaini, Jonathan Masci, Emanuele Rodol{\`a}, Michael~M Bronstein, and
  Daniel Cremers.
\newblock Anisotropic diffusion descriptors.
\newblock In {\em Computer Graphics Forum}, volume~35, pages 431--441. Wiley
  Online Library, 2016.

\bibitem{cohen2019gauge}
Taco Cohen, Maurice Weiler, Berkay Kicanaoglu, and Max Welling.
\newblock Gauge equivariant convolutional networks and the icosahedral cnn.
\newblock In {\em International Conference on Machine Learning}, pages
  1321--1330. PMLR, 2019.

\bibitem{de2020gauge}
Pim De~Haan, Maurice Weiler, Taco Cohen, and Max Welling.
\newblock Gauge equivariant mesh cnns: Anisotropic convolutions on geometric
  graphs.
\newblock In {\em International Conference on Learning Representations}, 2020.

\bibitem{shi2019neural}
Guanya Shi, Xichen Shi, Michael O’Connell, Rose Yu, Kamyar Azizzadenesheli,
  Animashree Anandkumar, Yisong Yue, and Soon-Jo Chung.
\newblock Neural lander: Stable drone landing control using learned dynamics.
\newblock In {\em 2019 International Conference on Robotics and Automation
  (ICRA)}, pages 9784--9790. IEEE, 2019.

\bibitem{zhong2019symplectic}
Yaofeng~Desmond Zhong, Biswadip Dey, and Amit Chakraborty.
\newblock Symplectic ode-net: Learning hamiltonian dynamics with control.
\newblock In {\em International Conference on Learning Representations}, 2020.

\bibitem{holl2019learning}
Philipp Holl, Nils Thuerey, and Vladlen Koltun.
\newblock Learning to control pdes with differentiable physics.
\newblock In {\em International Conference on Learning Representations}, 2019.

\bibitem{yin2021augmenting}
Yuan Yin, Vincent~LE GUEN, J{\'e}r{\'e}mie DONA, Emmanuel de~Bezenac, Ibrahim
  Ayed, Nicolas THOME, and patrick gallinari.
\newblock Augmenting physical models with deep networks for complex dynamics
  forecasting.
\newblock In {\em International Conference on Learning Representations}, 2021.

\bibitem{bronstein2021geometric}
Michael~M Bronstein, Joan Bruna, Taco Cohen, and Petar Veli{\v{c}}kovi{\'c}.
\newblock Geometric deep learning: Grids, groups, graphs, geodesics, and
  gauges.
\newblock {\em arXiv preprint arXiv:2104.13478}, 2021.

\bibitem{s.2018spherical}
Taco~S. Cohen, Mario Geiger, Jonas Köhler, and Max Welling.
\newblock Spherical {CNN}s.
\newblock In {\em International Conference on Learning Representations}, 2018.

\bibitem{zaheer2017deep}
Manzil Zaheer, Satwik Kottur, Siamak Ravanbakhsh, Barnabas Poczos, Russ~R
  Salakhutdinov, and Alexander~J Smola.
\newblock Deep sets.
\newblock {\em Advances in Neural Information Processing Systems}, 30, 2017.

\bibitem{long2018pde}
Zichao Long, Yiping Lu, Xianzhong Ma, and Bin Dong.
\newblock Pde-net: Learning pdes from data.
\newblock {\em International Conference on Machine Learning}, 2018.

\bibitem{long2019pde}
Zichao Long, Yiping Lu, and Bin Dong.
\newblock Pde-net 2.0: Learning pdes from data with a numeric-symbolic hybrid
  deep network.
\newblock {\em Journal of Computational Physics}, 399:108925, 2019.

\bibitem{wang2019learning}
Yufei Wang, Ziju Shen, Zichao Long, and Bin Dong.
\newblock Learning to discretize: solving 1d scalar conservation laws via deep
  reinforcement learning.
\newblock {\em arXiv preprint arXiv:1905.11079}, 2019.

\bibitem{xue2020amortized}
Tianju Xue, Alex Beatson, Sigrid Adriaenssens, and Ryan Adams.
\newblock Amortized finite element analysis for fast pde-constrained
  optimization.
\newblock In {\em International Conference on Machine Learning}, pages
  10638--10647. PMLR, 2020.

\bibitem{pfaff2020learning}
Tobias Pfaff, Meire Fortunato, Alvaro Sanchez-Gonzalez, and Peter Battaglia.
\newblock Learning mesh-based simulation with graph networks.
\newblock In {\em International Conference on Learning Representations}, 2020.

\bibitem{tremblay2018training}
Jonathan Tremblay, Aayush Prakash, David Acuna, Mark Brophy, Varun Jampani, Cem
  Anil, Thang To, Eric Cameracci, Shaad Boochoon, and Stan Birchfield.
\newblock Training deep networks with synthetic data: Bridging the reality gap
  by domain randomization.
\newblock In {\em Proceedings of the IEEE conference on computer vision and
  pattern recognition workshops}, pages 969--977, 2018.

\bibitem{bousmalis2018using}
Konstantinos Bousmalis, Alex Irpan, Paul Wohlhart, Yunfei Bai, Matthew Kelcey,
  Mrinal Kalakrishnan, Laura Downs, Julian Ibarz, Peter Pastor, Kurt Konolige,
  et~al.
\newblock Using simulation and domain adaptation to improve efficiency of deep
  robotic grasping.
\newblock In {\em 2018 IEEE international conference on robotics and automation
  (ICRA)}, pages 4243--4250. IEEE, 2018.

\bibitem{mueller2018driving}
Matthias Mueller, Alexey Dosovitskiy, Bernard Ghanem, and Vladlen Koltun.
\newblock Driving policy transfer via modularity and abstraction.
\newblock In {\em Conference on Robot Learning}, pages 1--15. PMLR, 2018.

\bibitem{jia2021physics}
Xiaowei Jia, Jared Willard, Anuj Karpatne, Jordan~S Read, Jacob~A Zwart,
  Michael Steinbach, and Vipin Kumar.
\newblock Physics-guided machine learning for scientific discovery: An
  application in simulating lake temperature profiles.
\newblock {\em ACM/IMS Transactions on Data Science}, 2(3):1--26, 2021.

\bibitem{garcia2019combining}
Victor Garcia~Satorras, Zeynep Akata, and Max Welling.
\newblock Combining generative and discriminative models for hybrid inference.
\newblock {\em Advances in Neural Information Processing Systems},
  32:13825--13835, 2019.

\bibitem{raissi2019physics}
Maziar Raissi, Paris Perdikaris, and George~E Karniadakis.
\newblock Physics-informed neural networks: A deep learning framework for
  solving forward and inverse problems involving nonlinear partial differential
  equations.
\newblock {\em Journal of Computational Physics}, 378:686--707, 2019.

\bibitem{Shi_Mo_Di_2021}
Rongye Shi, Zhaobin Mo, and Xuan Di.
\newblock Physics-informed deep learning for traffic state estimation: A hybrid
  paradigm informed by second-order traffic models.
\newblock {\em Proceedings of the AAAI Conference on Artificial Intelligence},
  35(1):540--547, May 2021.

\bibitem{yang2019enforcing}
Zeng Yang, Jin-Long Wu, and Heng Xiao.
\newblock Enforcing deterministic constraints on generative adversarial
  networks for emulating physical systems.
\newblock {\em arXiv preprint arXiv:1911.06671}, 2019.

\bibitem{wu2020enforcing}
Jin-Long Wu, Karthik Kashinath, Adrian Albert, Dragos Chirila, Heng Xiao,
  et~al.
\newblock Enforcing statistical constraints in generative adversarial networks
  for modeling chaotic dynamical systems.
\newblock {\em Journal of Computational Physics}, 406:109209, 2020.

\bibitem{Toth2020Hamiltonian}
Peter Toth, Danilo~J. Rezende, Andrew Jaegle, Sébastien Racanière, Aleksandar
  Botev, and Irina Higgins.
\newblock Hamiltonian generative networks.
\newblock In {\em International Conference on Learning Representations}, 2020.

\bibitem{satorras2021n}
Victor~Garcia Satorras, Emiel Hoogeboom, and Max Welling.
\newblock E (n) equivariant graph neural networks.
\newblock In {\em International Conference on Machine Learning}, 2021.

\bibitem{horie2021isometric}
Masanobu Horie, Naoki Morita, Toshiaki Hishinuma, Yu~Ihara, and Naoto Mitsume.
\newblock Isometric transformation invariant and equivariant graph
  convolutional networks.
\newblock In {\em International Conference on Learning Representations}, 2021.

\bibitem{weiler2019general}
Maurice Weiler and Gabriele Cesa.
\newblock General e (2)-equivariant steerable cnns.
\newblock {\em Advances in Neural Information Processing Systems},
  32:14334--14345, 2019.

\bibitem{thomas2018tensor}
Nathaniel Thomas, Tess Smidt, Steven Kearnes, Lusann Yang, Li~Li, Kai Kohlhoff,
  and Patrick Riley.
\newblock Tensor field networks: Rotation-and translation-equivariant neural
  networks for 3d point clouds.
\newblock {\em arXiv preprint arXiv:1802.08219}, 2018.

\bibitem{fuchs2020se}
Fabian Fuchs, Daniel Worrall, Volker Fischer, and Max Welling.
\newblock Se (3)-transformers: 3d roto-translation equivariant attention
  networks.
\newblock {\em Advances in Neural Information Processing Systems}, 33, 2020.

\bibitem{finzi2020generalizing}
Marc Finzi, Samuel Stanton, Pavel Izmailov, and Andrew~Gordon Wilson.
\newblock Generalizing convolutional neural networks for equivariance to lie
  groups on arbitrary continuous data.
\newblock In {\em International Conference on Machine Learning}, pages
  3165--3176. PMLR, 2020.

\bibitem{hutchinson2021lietransformer}
Michael~J Hutchinson, Charline Le~Lan, Sheheryar Zaidi, Emilien Dupont,
  Yee~Whye Teh, and Hyunjik Kim.
\newblock Lietransformer: Equivariant self-attention for lie groups.
\newblock In {\em International Conference on Machine Learning}, pages
  4533--4543. PMLR, 2021.

\bibitem{thanasutives2021adversarial}
Pongpisit Thanasutives, Ken-ichi Fukui, and Masayuki Numao.
\newblock Adversarial multi-task learning enhanced physics-informed neural
  networks for solving partial differential equations.
\newblock {\em arXiv preprint arXiv:2104.14320}, 2021.

\bibitem{seo2020physics}
Sungyong Seo, Chuizheng Meng, Sirisha Rambhatla, and Yan Liu.
\newblock Physics-aware spatiotemporal modules with auxiliary tasks for
  meta-learning.
\newblock In {\em IJCAI}, 2021.

\bibitem{jiang2018spherical}
Chiyu~Max Jiang, Jingwei Huang, Karthik Kashinath, Prabhat, Philip Marcus, and
  Matthias Niessner.
\newblock Spherical {CNN}s on unstructured grids.
\newblock In {\em International Conference on Learning Representations}, 2019.

\bibitem{alet2019graph}
Ferran Alet, Adarsh~Keshav Jeewajee, Maria~Bauza Villalonga, Alberto Rodriguez,
  Tomas Lozano-Perez, and Leslie Kaelbling.
\newblock Graph element networks: adaptive, structured computation and memory.
\newblock In {\em International Conference on Machine Learning}, pages
  212--222. PMLR, 2019.

\bibitem{trask2019gmls}
Nathaniel Trask, Ravi~G Patel, Ben~J Gross, and Paul~J Atzberger.
\newblock Gmls-nets: A framework for learning from unstructured data.
\newblock {\em arXiv preprint arXiv:1909.05371}, 2019.

\bibitem{long2018hybridnet}
Yun Long, Xueyuan She, and Saibal Mukhopadhyay.
\newblock Hybridnet: integrating model-based and data-driven learning to
  predict evolution of dynamical systems.
\newblock In {\em Conference on Robot Learning}, pages 551--560. PMLR, 2018.

\bibitem{takeishi2017learning}
Naoya Takeishi, Yoshinobu Kawahara, and Takehisa Yairi.
\newblock Learning koopman invariant subspaces for dynamic mode decomposition.
\newblock In {\em NIPS}, 2017.

\bibitem{li2019propagation}
Yunzhu Li, Jiajun Wu, Jun-Yan Zhu, Joshua~B Tenenbaum, Antonio Torralba, and
  Russ Tedrake.
\newblock Propagation networks for model-based control under partial
  observation.
\newblock In {\em 2019 International Conference on Robotics and Automation
  (ICRA)}, pages 1205--1211. IEEE, 2019.

\bibitem{finn2017model}
Chelsea Finn, Pieter Abbeel, and Sergey Levine.
\newblock Model-agnostic meta-learning for fast adaptation of deep networks.
\newblock In {\em Proceedings of the 34th International Conference on Machine
  Learning-Volume 70}, pages 1126--1135. JMLR. org, 2017.

\bibitem{shu1998essentially}
Chi-Wang Shu.
\newblock Essentially non-oscillatory and weighted essentially non-oscillatory
  schemes for hyperbolic conservation laws.
\newblock In {\em Advanced numerical approximation of nonlinear hyperbolic
  equations}, pages 325--432. Springer, 1998.

\bibitem{lim2007mls}
Jae~Hyuk Lim, Seyoung Im, and Young-Sam Cho.
\newblock Mls (moving least square)-based finite elements for three-dimensional
  nonmatching meshes and adaptive mesh refinement.
\newblock {\em Computer methods in applied mechanics and engineering},
  196(17-20):2216--2228, 2007.

\bibitem{feynman2005principle}
Richard~P Feynman.
\newblock The principle of least action in quantum mechanics.
\newblock In {\em Feynman's Thesis—A New Approach To Quantum Theory}, pages
  1--69. World Scientific, 2005.

\bibitem{elsken2019neural}
Thomas Elsken, Jan~Hendrik Metzen, and Frank Hutter.
\newblock Neural architecture search: A survey.
\newblock {\em The Journal of Machine Learning Research}, 20(1):1997--2017,
  2019.

\bibitem{skomski2021automating}
Elliott Skomski, J{\'a}n Drgo{\v{n}}a, and Aaron Tuor.
\newblock Automating discovery of physics-informed neural state space models
  via learning and evolution.
\newblock In {\em Learning for Dynamics and Control}, pages 980--991. PMLR,
  2021.

\bibitem{alet2019neural}
Ferran Alet, Erica Weng, Tom{\'a}s Lozano-P{\'e}rez, and Leslie~Pack Kaelbling.
\newblock Neural relational inference with fast modular meta-learning.
\newblock In {\em Advances in Neural Information Processing Systems}, pages
  11804--11815, 2019.

\bibitem{chen2020modular}
Yutian Chen, Abram~L Friesen, Feryal Behbahani, Arnaud Doucet, David Budden,
  Matthew Hoffman, and Nando de~Freitas.
\newblock Modular meta-learning with shrinkage.
\newblock {\em Advances in Neural Information Processing Systems},
  33:2858--2869, 2020.

\bibitem{goyal2021recurrent}
Anirudh Goyal, Alex Lamb, Jordan Hoffmann, Shagun Sodhani, Sergey Levine,
  Yoshua Bengio, and Bernhard Sch{\"o}lkopf.
\newblock Recurrent independent mechanisms.
\newblock In {\em International Conference on Learning Representations}, 2021.

\bibitem{ILSVRC15}
Olga Russakovsky, Jia Deng, Hao Su, Jonathan Krause, Sanjeev Satheesh, Sean Ma,
  Zhiheng Huang, Andrej Karpathy, Aditya Khosla, Michael Bernstein,
  Alexander~C. Berg, and Li~Fei-Fei.
\newblock {ImageNet Large Scale Visual Recognition Challenge}.
\newblock {\em International Journal of Computer Vision (IJCV)},
  115(3):211--252, 2015.

\bibitem{lin2014microsoft}
Tsung-Yi Lin, Michael Maire, Serge Belongie, James Hays, Pietro Perona, Deva
  Ramanan, Piotr Doll{\'a}r, and C~Lawrence Zitnick.
\newblock Microsoft coco: Common objects in context.
\newblock In {\em European conference on computer vision}, pages 740--755.
  Springer, 2014.

\bibitem{bojar2014findings}
Ond{\v{r}}ej Bojar, Christian Buck, Christian Federmann, Barry Haddow, Philipp
  Koehn, Johannes Leveling, Christof Monz, Pavel Pecina, Matt Post, Herve
  Saint-Amand, et~al.
\newblock Findings of the 2014 workshop on statistical machine translation.
\newblock In {\em Proceedings of the ninth workshop on statistical machine
  translation}, pages 12--58, 2014.

\bibitem{rajpurkar2016squad}
Pranav Rajpurkar, Jian Zhang, Konstantin Lopyrev, and Percy Liang.
\newblock Squad: 100, 000+ questions for machine comprehension of text.
\newblock In {\em EMNLP}, 2016.

\bibitem{hu2020open}
Weihua Hu, Matthias Fey, Marinka Zitnik, Yuxiao Dong, Hongyu Ren, Bowen Liu,
  Michele Catasta, and Jure Leskovec.
\newblock Open graph benchmark: Datasets for machine learning on graphs.
\newblock {\em arXiv preprint arXiv:2005.00687}, 2020.

\bibitem{he2015delving}
Kaiming He, Xiangyu Zhang, Shaoqing Ren, and Jian Sun.
\newblock Delving deep into rectifiers: Surpassing human-level performance on
  imagenet classification.
\newblock In {\em Proceedings of the IEEE international conference on computer
  vision}, pages 1026--1034, 2015.

\bibitem{glorot2010understanding}
Xavier Glorot and Yoshua Bengio.
\newblock Understanding the difficulty of training deep feedforward neural
  networks.
\newblock In {\em Proceedings of the thirteenth international conference on
  artificial intelligence and statistics}, pages 249--256. JMLR Workshop and
  Conference Proceedings, 2010.

\bibitem{jagtap2020adaptive}
Ameya~D Jagtap, Kenji Kawaguchi, and George~Em Karniadakis.
\newblock Adaptive activation functions accelerate convergence in deep and
  physics-informed neural networks.
\newblock {\em Journal of Computational Physics}, 404:109136, 2020.

\bibitem{sitzmann2020implicit}
Vincent Sitzmann, Julien Martel, Alexander Bergman, David Lindell, and Gordon
  Wetzstein.
\newblock Implicit neural representations with periodic activation functions.
\newblock {\em Advances in Neural Information Processing Systems}, 33, 2020.

\bibitem{kim2021dpm}
Jungeun Kim, Kookjin Lee, Dongeun Lee, Sheo~Yon Jhin, and Noseong Park.
\newblock Dpm: A novel training method for physics-informed neural networks in
  extrapolation.
\newblock In {\em Proceedings of the AAAI Conference on Artificial
  Intelligence}, volume~35, pages 8146--8154, 2021.

\end{thebibliography}

\end{document}